\documentclass[10pt,twocolumn,letterpaper]{article}

\usepackage{iccv}              

\definecolor{iccvblue}{rgb}{0.21,0.49,0.74}
\usepackage[pagebackref,breaklinks,colorlinks,allcolors=iccvblue]{hyperref}
\usepackage{colortbl}
\usepackage{float}
\usepackage{placeins}
\usepackage{multirow}
\usepackage{algorithm}
\usepackage{algorithmic}
\def\paperID{10908} 
\def\confName{ICCV}
\def\confYear{2025}
\DeclareUnicodeCharacter{FF5E}{\textasciitilde}

\title{IMPACT: Behavioral \underline{I}ntention-aware \underline{M}ultimodal Trajectory \underline{P}rediction with \underline{A}daptive \underline{C}ontext \underline{T}rimming}

\author{
Jiawei Sun$^{1}$\thanks{This work was done during his internship at Xiaomi EV.}, Xibin Yue$^{2}$, Jiahui Li$^{1}$, Tianle Shen$^{1}$, Chengran Yuan$^{1}$,\\
Shuo Sun$^{1}$, Sheng Guo$^{1}$, Quanyun Zhou$^{2}$, Marcelo H. Ang Jr$^{1}$ \\
$^{1}$National University of Singapore \hspace{2em} $^{2}$Xiaomi EV 
}
\begin{document}
\maketitle
\begin{abstract}
While most prior research has focused on improving the precision of multimodal trajectory predictions, the explicit modeling of multimodal behavioral intentions (e.g., yielding, overtaking) remains relatively underexplored. This paper proposes a unified framework that jointly predicts both behavioral intentions and trajectories to enhance prediction accuracy, interpretability, and efficiency. Specifically, we employ a shared context encoder for both intention and trajectory predictions, thereby reducing structural redundancy and information loss. Moreover, we address the lack of ground-truth behavioral intention labels in mainstream datasets (Waymo, Argoverse) by auto-labeling these datasets, thus advancing the community’s efforts in this direction. We further introduce a vectorized occupancy prediction module that infers the probability of each map polyline being occupied by the target vehicle’s future trajectory. By leveraging these intention and occupancy predictions priors, our method conducts dynamic, modality-dependent pruning of irrelevant agents and map polylines in the decoding stage, effectively reducing computational overhead and mitigating noise from non-critical elements. 
Our approach ranks first among LiDAR-free methods on the Waymo Motion Dataset and achieves first place on the Waymo Interactive Prediction Dataset. Remarkably, even without model ensembling, our single-model framework improves the softmAP by 10\% compared to the second-best method in Waymo Interactive Prediction Leaderboard. Furthermore, the proposed framework has been successfully deployed on real vehicles, demonstrating its practical effectiveness in real-world applications.
\end{abstract}    

\section{Introduction}

In the realm of autonomous driving, accurately predicting the future behaviors of surrounding agents is crucial to ensure safe, efficient, and comfortable driving experience. 

\begin{figure}
    \centering
    \setlength{\abovecaptionskip}{-0.2cm}
        \includegraphics[width=1\linewidth]{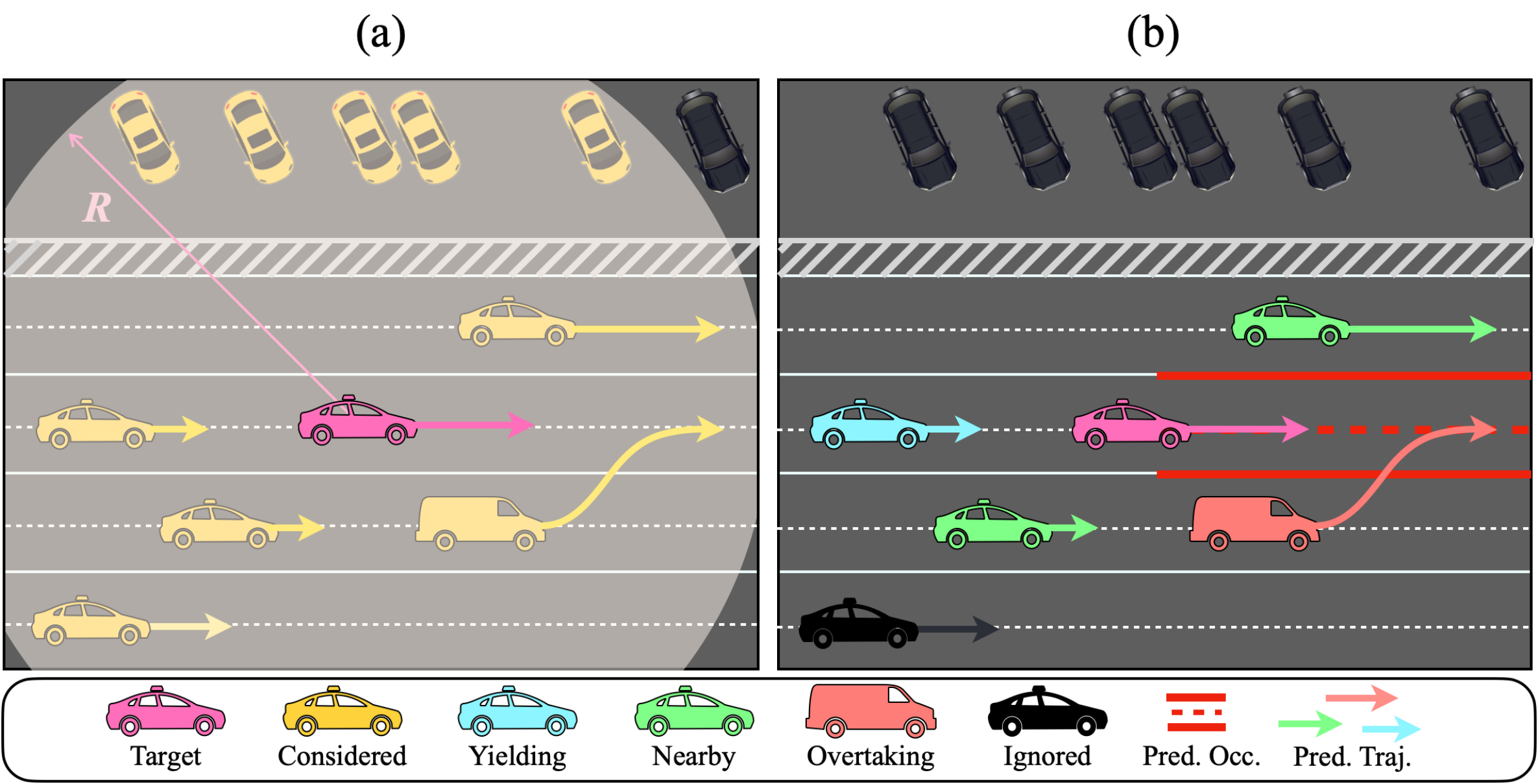}
    \hfill
    \caption{(a) illustrates the traditional context input, while (b) is our integrated approach jointly predicting behavioral intentions, trajectories, and vectorized occupancy. In our approach, the decoder stage is fed only with influential agents and relevant map elements. }
    \label{figure-cover}
    \vspace{-4mm}
\end{figure}
With mainstream approaches including \cite{wang2024futurenetlofjointtrajectoryprediction, zhou2024modeseq, shi2024mtr++, Zhou_2023_qc} prioritising generating precise multimodal trajectory predictions, most recent research focused on achieving higher average precisions (mAP) and minimizing deviations (Brier-minFDE). However, humans primarily determine the future behavior of interacting agents not by focusing on specific states alone, but rather by inferring the underlying intention that drives their actions.

This insight is also applicable in autonomous driving, where predicting surrounding agents' future poses alone is insufficient and partial; understanding their behavioral intention (eg, yielding or overtaking) toward the ego vehicle is equally critical. Without explicit intention information, the downstream decision-making and planning modules are left with only a distribution of possible future coordinates. 
They must then infer the other vehicle’s underlying behavior in an additional step, which not only complicates the ego vehicle’s decision-making process but can also reduce the overall reliability and responsiveness of the system. 
The most straightforward method is to additionally incorporate one more behavioral intention prediction module, but this decoupled architecture inevitably introduces structural redundancy and information fragmentation. 

Currently, most trajectory prediction models use attention mechanism for querying information from agents and map features. However, indiscriminately attending to all agents and map elements introduces unnecessary complexity and potential causal confusion, as many elements have no direct relevance to the target agent, ultimately hindering correct decoding of the target agent’s future trajectories. To address this, some recent approaches introduce an additional local context-aware refinement module after the decoder stage \cite{zhou2024smartrefinescenarioadaptiverefinementframework, choi2023rpred}, but this further complicates the prediction pipeline. Others adopt rule-based heuristic pruning of input lanes \cite{lan2023SEPT} or dynamically collect map elements based on the previous layer prediction results \cite{shi2023MTR, shi2024mtr++, lin2024eda, sun2024rmpyolorobustmotionpredictor}. However, these methods may struggle in complex scenarios or suffer from error propagation. BETOP \cite{liu2024reasoningmultiagentbehavioraltopology} is the first paper explicitly modelling agent interactions through braid theory, but its topological selection algorithm demonstrates critical accuracy deficiencies (see appendix \autoref{fig:comparebetop}).  Thus, an efficient approach to selectively focus on truly critical features (both agents and maps) remains an open challenge. 
Furthermore, the lack of ground-truth intention labels in current mainstream datasets (e.g., Waymo \cite{waymo_motion_prediction}, Argoverse \cite{argoverse,argoverse2}) is limiting the development of accurate behavioral intention-aware models.

To holistically address the aforementioned challenges, we propose our approach, IMPACT. Specifically, we introduce an agent-wise multimodal behavioral intention module before the trajectory decoder, while sharing the same scenario information extraction module (i.e., the same encoder) used by the trajectory predictor. This design reduces computational overhead and mitigates information loss. The agent-wise behavioral intention module outputs a one-hot intention vector for each vehicle, where each element in the vector represents the probability of a specific intention relative to the target vehicle (e.g., overtaking, yielding). Based on these probability vectors, we apply a utility function to assign an “interaction score” to each vehicle, selecting only those with high interaction probabilities as inputs to the decoder. In parallel, we propose a vectorized occupancy prediction module. This module predicts the probability that each vectorized map element will be occupied by the target vehicle in the future. Similar to the agent selection process, we only feed the map polylines with high occupancy probabilities into the decoder. By integrating both modules in this manner, our method effectively reduces redundant interactions while preserving crucial information for accurate and robust trajectory prediction. We also propose an automatic labeling algorithm to generate high-quality ground truth labels for behavioral intentions on the Waymo and Argoverse 1\&2 dataset.

We evaluated IMPACT on both the Waymo Motion Prediction Dataset (for marginal prediction) and the Waymo Interactive Prediction Dataset (for interaction prediction). Experimental results demonstrate that our method achieves the best performance among all LiDAR-free methods on the Waymo Motion Prediction Dataset, surpassed only by MTRV3 \cite{shi2024mtrv3}, which leverages additional LiDAR data. Moreover, IMPACT ranks first overall on the Waymo Interactive Prediction Dataset, surpassing the previous state-of-the-art ensemble model \cite{liu2024reasoningmultiagentbehavioraltopology} using only a single model. To further demonstrate the practical applicability of our approach, we successfully trained IMPACT on a proprietary dataset and deployed it effectively on a real vehicle.

Our main contributions can be summarized as follows:

1) \textbf{Intent-Integrated Trajectory Prediction.}  
We jointly predict agent multimodal intentions and trajectories in a unified framework, eliminating redundant modules and enhancing information flow.

2) \textbf{Context-Aware Pruning via Dual Filters.}  
We introduce complementary agent and map filters that leverage predicted behavioral interaction probabilities and vectorized occupancy to retain only influential vehicles and relevant map elements.

3) \textbf{Automatic Intention Label Generation.}  
We propose an automatic labeling strategy to annotate agent-level intentions in large-scale datasets (e.g., Waymo, Argoverse). This strategy enables more convenient behavior prediction without manual effort.

4) \textbf{State-of-the-Art Performance and Real-World Validation.}  
IMPACT achieves SOTA results on mainstream public benchmarks and demonstrates robust real-world performance through deployment on an autonomous vehicle (a video demo in supplementary material). \\

\section{Related Works}
\label{sec:Related Works}

\subsection{Motion Prediction}
Early approaches \cite{chai2019multipathmultipleprobabilisticanchor,gilles2021homeheatmapoutputfuture,phanminh2020covernetmultimodalbehaviorprediction} relied on Bird's-Eye-View (BEV) rasterized representations processed through CNNs, but struggled to preserve geometric fidelity and semantic relationships. This limitation spurred vectorized paradigms like VectorNet \cite{gao2020vectornetencodinghdmaps}, which encoded agents and map elements as polylines. Subsequent graph-based architectures \cite{deo2021multimodaltrajectorypredictionconditioned, JiaHDGT} further formalized relational dependencies through graph neural networks. The field then diverged into anchor-oriented strategies \cite{zhou2024smartrefinescenarioadaptiverefinementframework,MultiPath++, gu2021densetnt,zhao2020tnt} leveraging HD map anchors for physically constrained prediction, followed by agent-centric frameworks \cite{Hivt,shi2023MTR}. But it requires re-normalizing and re-encoding during online inference. Query-centric approaches\cite{zhou2023qcnext, Zhou_2023_qc, zhang2024simple, shi2024mtr++, futurelof} addressed this by encoding each vehicle as a query, leveraging attention mechanisms to enhance prediction performance and improve multi-agent coherence through unified spatio-temporal encoding. Recent research in motion forecasting has witnessed significant advancements across multiple frontiers, driving improvements in generalization, consistency, and predictive accuracy. Self-supervised approaches \cite{bhattacharyya2022ssllane,forecastmae,lan2023SEPT} enhance model generalization by leveraging reconstruction pretext tasks, while temporal consistency methods \cite{zhang2024demodecouplingmotionforecasting,tang2024hpnet,realmotion} reinforce scene coherence through continuous-time modeling. At the same time, architectures like RMP-YOLO \cite{sun2024rmpyolorobustmotionpredictor,POP_RAL} tackle the challenge of partial observations by imputing missing trajectory segments. Furthermore, ModeSeq \cite{zhou2024modeseq} introduces a paradigm shift by framing motion prediction as a GPT-style next-token prediction task.

However, current methods tend to emphasize the precise generation of multimodal trajectory predictions at the expense of behavioral intention prediction\cite{fang2023behavioralintentionpredictiondriving} (BIP). This narrow focus often overlooks the latent decision-making processes that drive observable maneuvers, resulting in models that struggle to interpret complex social interactions or anticipate nuanced changes in driving behavior. In real-world scenarios, the inability to infer intentions such as overtake or yield decisions can lead to suboptimal planning and reduced overall robustness. 
The most recent work, BeTOP~\cite{liu2024reasoningmultiagentbehavioraltopology}, attempts to explicitly model behavioral interactions via braid theory by splitting agents into “interactive” and “non-interactive” based on whether their future trajectories form an intertwined braid. However, in most scenarios (see Appendix \autoref{fig:comparebetop}), the generated braids appear unrealistic because the approach only considers lateral and temporal dimensions. Moreover, relying on a binary behavioral label oversimplifies complex traffic scenarios, where interactions can be far more nuanced.

Our IMPACT framework addresses these limitations by jointly predicting multimodal, multi-class behavioral intentions and future trajectories. To better supervise the behavioral intention predictor, we propose an auto-labeling algorithm that generates reasonable intention labels, striking a balance between explainability and behavioral completeness.

\subsection{Context-aware Pruning}
During the decoder stage, current works often relies on attention mechanisms~\cite{vaswani2023attentionneed} to query agent-map information. However, using global attention over all entities incurs \(O(n^2)\) complexity and can introduce causal confusion, in contrast to human drivers, who primarily focus on goal-critical paths and potentially interactive vehicles~\cite{fang2023behavioralintentionpredictiondriving}. Some approaches address this via prior rule-based pruning (e.g., SEPT~\cite{lan2023SEPT}), which may fail for corner case. Meanwhile, MTR-series\cite{shi2023MTR,shi2024mtr++,sun2024controlmtr,lin2023eda,mu2024mostmultimodalityscenetokenization,gan2024mgtrmultigranulartransformermotion,sun2024rmpyolorobustmotionpredictor,llm_augmented_mtr} methods and R-pred\cite{choi2023rpred} use last layer predictions as priors to dynamically collect nearby polylines, yet suffers from error propogation. BeTOP\cite{liu2024reasoningmultiagentbehavioraltopology}, which relies on braid theory to select interactive vehicles, can yield an unreasonably chosen set of vehicles in practice. To tackle these issues, our IMPACT framework introduces a symmetric dual context-filtering approach that leverage predicted behavioral interaction probabilities and vectorized 
occupancy to retain only influential vehicles and relevant map elements.

\section{Methodologies}
\label{sec:Methodologies}

\subsection{Problem Formulation}
Following the vectorized representation in VectorNet, we denote the historical trajectories of $N_a$ traffic participants as $\mathcal{A}=\{a_{1}, a_{2},\dots, a_{N_{a}}\}$. The corresponding map is equally partitioned into $N_{l}$ polylines $\mathcal{L}=\{l_{1},l_{2},\dots,l_{N_{l}}\}$. The predictor will anticipate $K$ different modality future trajectories $\mathcal{Y}=\{y_{1},y_{2},\dots,y_{N_{a}}\}$ over the future $T_{f}$ timesteps, where $y_{i}=\{y_{i}^{1},y_{i}^{2},\dots,y_{i}^{K}\}\in \mathbb{R}^{ K \times T_{f} \times 2}$. The confidence score for $y_{i}$ are denoted as $s_{i} = \{s_{i}^{1},s_{i}^{2},\dots,s_{i}^{K}\}$.
Then for target agent $a_{i}$, existing motion prediction task aims to estimate the distribution: 

\begingroup
\small
\begin{equation}
P(y_{i} \mid \mathcal{L}, \mathcal{A}) = \sum_{k=1}^{K} s_i^k P(y_{i}^{k} \mid \mathcal{L},\mathcal{A})
\end{equation}
\endgroup

To better capture inter-agent interactions and crucial map segments, we additional predict two sets of modality-dependent priors: behavioral intentions  $\mathcal{H}^k =\{h_1^k,\dots,h_{N_a}^k\},$ where $h_i^k\in\mathbb{R}^4$ encodes the probability distribution of behavioral intentions (overtaking, yielding, ignored, nearby) for agent $a_i$ toward target agent under mode $k$ and vectorized occupancy $ \mathcal{O}^k = \{o_1^k,\dots,o_{N_\ell}^k\},$
where $o_j^k \in [0,1]$ indicates the probability that polyline $l_j$ is relevant or “occupied” by the agent’s future path under mode $k$.
Given $\{\mathcal{H}^k\}$ and $\{\mathcal{O}^k\}$, we perform a top-$m$ selection of agents and top-$n$ selection of map polylines for each modality $k$:
\begingroup
\small
\begin{align}
\begin{split}    
    \mathcal{A}_k^{\mathrm{sel}} 
    = \Bigl\{
        a_j :\; j \in \mathrm{argtop}_{m}\bigl[\psi(h_j^k)\bigr]_{j=1}^{N_a} 
    \Bigr\},
    \\
    \mathcal{L}_k^{\mathrm{sel}} 
    = \Bigl\{
        l_j :\; j \in \mathrm{argtop}_{n}\bigl[\varphi(o_j^k)\bigr]_{j=1}^{N_\ell} 
    \Bigr\}.\
\end{split}
\end{align}
\endgroup
where $\psi(h_j^k)$ and $\varphi(\mathbf{o}_j^k)$ is a scalar score derived from the intention vector $\mathbf{h}_j^k$  and occupancy vector $\mathbf{o}_j^k$ by utility function $\psi(\cdot)$ and $\varphi(\cdot)$.
These subsets ensure that each modality $k$ focuses only on the most critical agents and polylines.

To jointly capture both the behavioral intentions and the map occupancy, we consider the extended distribution:
\begingroup
\small
\begin{equation}
\begin{aligned}
&P\Bigl(y_i,\;\mathcal{H},\;\mathcal{O} \;\Big|\; \mathcal{L}, \mathcal{A}\Bigr)
\approx\\ 
&\sum_{k=1}^{K}
s_{i}^{k}P\bigl(\mathcal{H}^k \mid \mathcal{L}, \mathcal{A}\bigr)\,
P\bigl(\mathcal{O}^k \mid \mathcal{L}, \mathcal{A}\bigr)\;
P\Bigl(y_i^{k} \mid 
\mathcal{A}_k^{\mathrm{sel}},\,
\mathcal{L}_k^{\mathrm{sel}}\Bigr).
\end{aligned}
\end{equation}
\endgroup

\begin{figure*}
    \centering
    \setlength{\abovecaptionskip}{-0.2cm}
    \includegraphics[width=16cm]{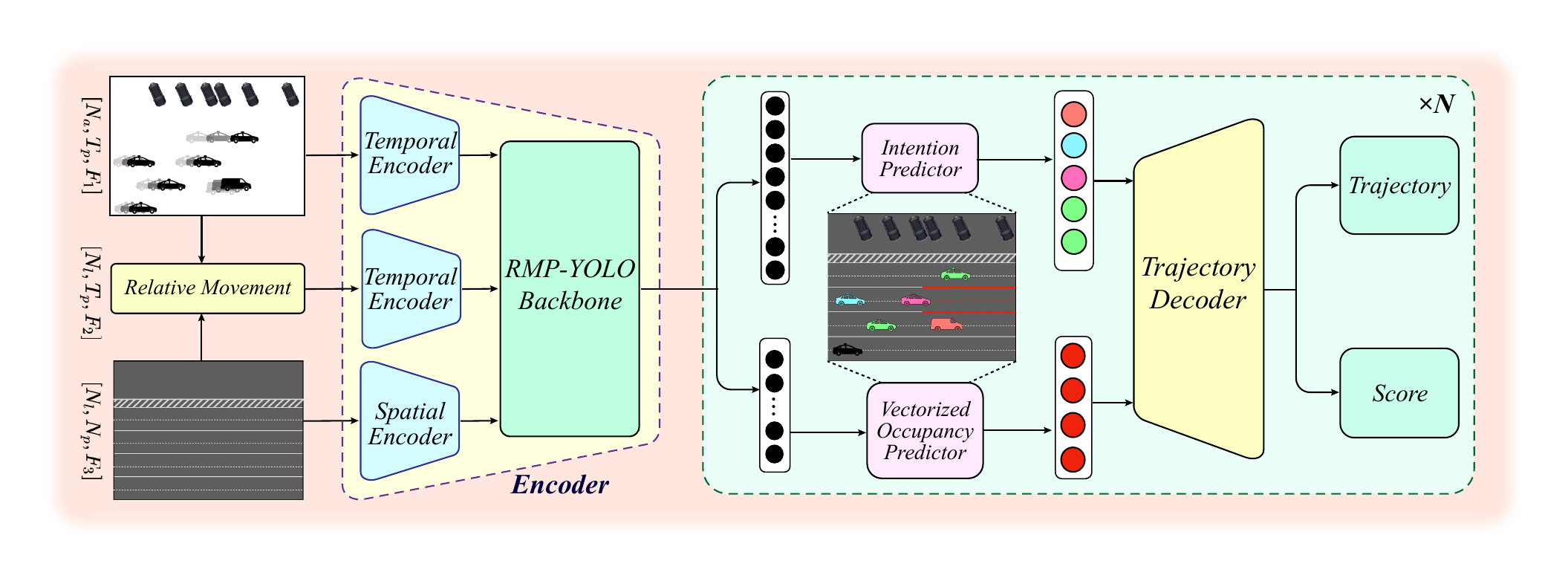} 
    \caption{An overview of framework of IMPACT. Both the Intention Predictor and the Vectorized Occupancy Predictor share the same context encoder with the Trajectory Decoder, leveraging their outputs to prune irrelevant agents and map polylines. This selective mechanism ensures that only the most critical context is fed into the decoder for final trajectory prediction.}
    \label{fig:encoder-structure}
    \vspace{-1em}
\end{figure*}

Where $P\bigl(\mathcal{H}^k \mid \mathcal{L}, \mathcal{A}\bigr)$ yields the intention vectors for each agent under mode $k$, $P\bigl(\mathcal{O}^k \mid \mathcal{L}, \mathcal{A}\bigr)$ produces occupancy scores for each polyline under mode $k$, $\mathcal{A}_k^{\mathrm{sel}}$ and $\mathcal{L}_k^{\mathrm{sel}}$ are the selected agents/polylines for each modality, and
the final trajectory $y_i^k$ is decoded by attending only to these subsets. This operation reduces cross-attention complexity from $O(KN_a + KN_\ell)$ to $O(Km+Kn)$ while maintaining accuracy.

\subsection{Input Representation}
In our method, we apply agent-centric normalization. To predict a target agent, the input to the predictor comprises:
\( A=\{a_{1},a_{2},\dots,a_{N_{a}}\}\in \mathbb{R}^{N_{a} \times T_{p} \times F_{1}} \), representing \( N_a \) agents with \( T_p \) past states and feature dimension \( F_1 \), and
\( \mathcal{L}=\{l_{1},l_{2},\dots,l_{N_{l}}\} \in \mathbb{R}^{N_{l} \times N_{p} \times F_{2}} \), representing \( N_l \) polylines with \( N_p \) points each and feature dimension \( F_2 \).

Following the approaches of ControlMTR\cite{sun2024controlmtr}, RMP-YOLO\cite{sun2024rmpyolorobustmotionpredictor}, and MacFormer\cite{Feng_2023_macformer}, we further incorporate the historical relative movement between the target agent and the map polylines to capture dynamic subtle interdependencies. This historical movement is denoted by $
\mathcal{R}=\{r_{1},r_{2},\dots,r_{N_{l}}\} \in \mathbb{R}^{N_{l} \times T_{p} \times F_{3}},
$ where \( F_{3} \) is the feature size associated with the relative movement($(\Delta x,\, \Delta y,\, \cos\Delta\theta,\, \sin\Delta\theta)$).
\subsection{Network Structure}
\subsubsection{Spatial Temporal Encoding} To comprehensively model temporal dependencies, we apply a Multi-Scale LSTM (MSL) module. The time-series data \(\mathcal{A}\) and \(\mathcal{R}\) are each processed through three parallel streams. Each stream consists of a 1D CNN with a distinct kernel size followed by an LSTM. For the \(i\)-th stream with kernel size \(k_i\), the output at the final time step \(T_p\) is computed as: \

\begingroup
\small
\begin{align}
\mathcal{A}^{T_p}_{k_i} &= \text{LSTM}\left(\text{Conv1D}_{k_i}(\mathcal{A})\right)\Big|_{t=T_p}, \quad k_i \in \{1, 3, 5\} \\
\mathcal{R}^{T_p}_{k_i} &= \text{LSTM}\left(\text{Conv1D}_{k_i}(\mathcal{R})\right)\Big|_{t=T_p}, \quad k_i \in \{1, 3, 5\}
\end{align}
\endgroup
The final-step hidden states are concatenated ($\oplus$) along the feature dimension, and a multi-layer perceptron (MLP) projects the concatenated vector into a unified temporal feature token:
\begingroup
\small
\begin{align}
\mathcal{A}^1 = MLP(\mathcal{A}^{T_p}_{1}\oplus\mathcal{A}^{T_p}_{3}\oplus\mathcal{A}^{T_p}_{5})\in \mathbb{R}^{N_{a}\times D}, \\
\mathcal{R}^1 = MLP(\mathcal{R}^{T_p}_{1}\oplus\mathcal{R}^{T_p}_{3}\oplus\mathcal{R}^{T_p}_{5})\in \mathbb{R}^{N_{l}\times D}.
\end{align}
\endgroup
For the spatial data \(\mathcal{L}\), we adopt a simplified PointNet-like architecture to aggregate each polyline into a feature token: 
\begingroup
\small
\begin{equation}
    \mathcal{L}_1 = \mathrm{MaxPooling}(\mathrm{MLP}(\mathcal{L})) \in \mathbb{R}^{N_{l}\times D}.
\end{equation}
\endgroup

These tokens are aligned in the feature space (\(\mathbb{R}^D\)) and ready for downstream fusion and prediction tasks.

\subsubsection{Feature Fusion.} We utilize RMP-YOLO's encoder\cite{sun2024rmpyolorobustmotionpredictor} as a backbone network (see \autoref{fig:encoder-structure}) to do feature fusion. To integrate heterogeneous input modalities, we employ a cascaded Multi-Context Gating (MCG) mechanism inspired by \cite{MultiPath++}. The MCG modules sequentially fuse pairs of modalities from a candidate set of three. The output of each MCG stage serves as input to the subsequent stage, enabling hierarchical feature interaction:
\begingroup
\small
\begin{align}
(\mathcal{A}^{2}, \mathcal{R}^{2}) &= \mathrm{MCG}(\mathcal{A}^{1}, \mathcal{R}^{1}), \\
(\mathcal{L}^{2}, \mathcal{R}^{3}) &= \mathrm{MCG}(\mathcal{L}^{1}, \mathcal{R}^{2}), \\
(\mathcal{A}^{3}, \mathcal{L}^{3}) &= \mathrm{MCG}(\mathcal{A}^{2}, \mathcal{L}^{2}),
\end{align}
\endgroup

The final fused tokens are defined as  
agent tokens: $\mathcal{A}_{3} \in \mathbb{R}^{N_{a}\times D}$  and map tokens: $\mathcal{L}_{3} = \mathcal{L}_{3} + \mathcal{R}_{3} \in \mathbb{R}^{N_{p}\times D}$.
Thus, we design a K-nearest-neighbor (KNN) guided local attention mechanism to restrict each agent token to attend only to its $K$ most relevant neighboring tokens (agents or map elements). This sparse attention pattern reduces computational complexity while preserving critical interactions. Six transformer encoder layers are applied to achieve deep feature fusion. Each layer follows the standard transformer architecture enhanced with positional encoding:  
\begingroup
\small
\begin{equation}
    \begin{split}
        X^{i} = \mathrm{MHA} (
            & X^{i-1}+\mathrm{PE}({X^{i-1}}), \\ 
            & \mathcal{K}(X^{i-1}) + \mathrm{PE}({\mathcal{K}(X^{i-1}))}, 
            \mathcal{K}(X^{i-1}))
    \end{split}
\end{equation}
\endgroup

\begin{figure*}
    \centering
    \setlength{\abovecaptionskip}{-0.2cm}
    \includegraphics[width=18cm]{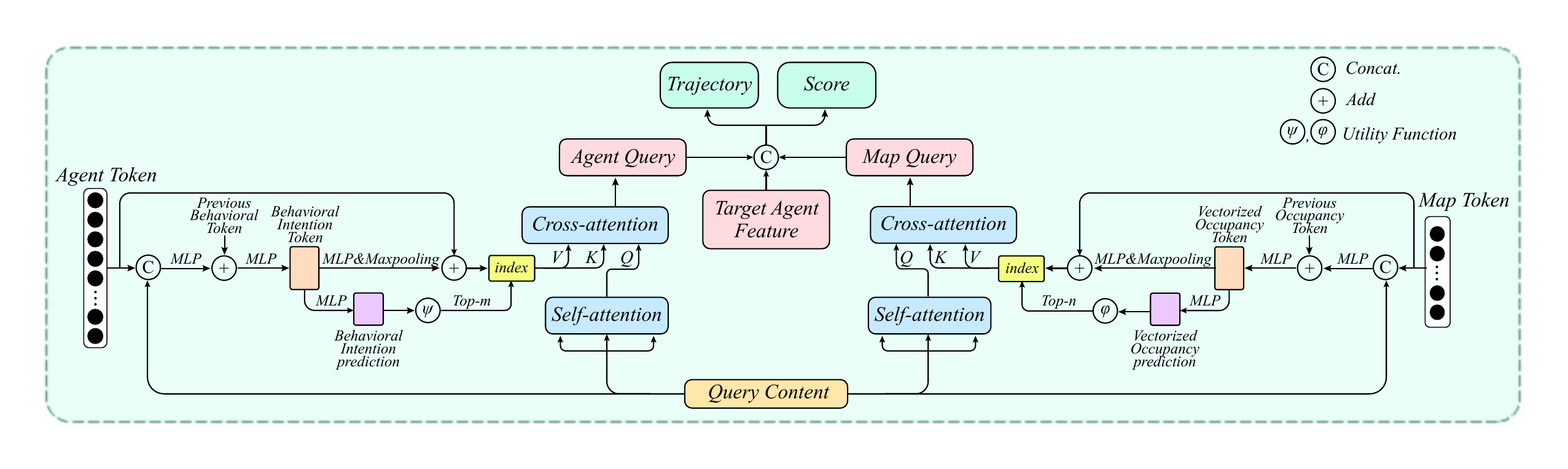} 
    \caption{An overview of our decoder framework, featuring context-aware pruning via symmetric dual filters.}
    \label{fig:decoder-structure}
    \vspace{-1em}
\end{figure*}where $X^{0} = [\mathcal{L}_{3},\, \mathcal{A}_{3}] \in \mathbb{R}^{(N_{a}+N_{m})\times D}$, $\mathrm{MHA}(\cdot)$ denotes multi-head attention, $\mathcal{K}(\cdot)$ selects $K$-nearest neighbors via Euclidean distance, and $\mathrm{PE}(\cdot)$ injects positional information using sinusoidal encoding. The positional coordinates derive from agents’ latest observed positions and map polylines’ centroid coordinates. The final output tokens $X^{Final} = [\mathcal{L}_{4},\, \mathcal{A}_{4}]$ are fed into the behavioral intention prediction module and vectorized occupation prediction for dynamic context-aware pruning.

Before diving into decoder part (see \autoref{fig:decoder-structure}), we define query content feature at decoder layer $i$ as ${Q}^{i}\in\mathbb{R}^{K\times D}$, which are later used to aggregate information from agent tokens and map tokens, and decode multimodal prediction results and $K$ denotes the number of different futures.

\subsubsection{Multimodal Behavioral Intention Prediction} 
For each future modality, we predict the behavioral intentions of other agents with respect to the target agent. Given the input agent tokens \(\mathcal{A}_4 \in \mathbb{R}^{N_a \times D}\) and the query content \(Q^{i} \in \mathbb{R}^{K \times D}\), we fuse these features into a unified representation of shape \(\mathbb{R}^{K \times N_a \times 2D}\) via straightforward tensor broadcasting and concatenation. Next, the fused features are passed through a multi-layer perceptron (MLP) and then added to the previous layer's behavioral intention token \(I^{i-1} \in \mathbb{R}^{K \times N_a \times D}\). Finally, another MLP followed by a softmax activation function produces the final behavioral intention predictions:
\begingroup
\small
\begin{equation}  
\begin{aligned}
\hat{\mathrm{H}^{i}} 
&= \text{Softmax}\Bigl(\text{MLP}\bigl(I^{i}\bigr)\Bigr) 
\;\in\; \mathbb{R}^{K \times N_a \times 4}, \\
I^{i} &= \text{MLP}(\text{MLP}(\mathcal{A}_4 \oplus Q^{i}) + I^{i-1}).
\end{aligned}
\end{equation}
\endgroup

Each vector element \(h\) represents a probability distribution over four intention categories: \textit{yielding}, \textit{overtaking}, \textit{ignored}, and \textit{nearby}. To make the decoding process more focused, we first compute an overall interaction score from the predicted distribution \(\hat{\mathrm{H}}\) using a utility function \(\psi\), producing \(\psi(\hat{\mathrm{H}^{i}})\in \mathbb{R}^{K\times N_a}\). We then select the top-\(m\) highest-scoring agents for downstream trajectory decoding. Therefore, for each modality, we choose $m$ most revelant agents \(A_{5}\in \mathbb{R}^{m\times D}\), where \(m \ll N_{a}\). This filtering step refines the decoder’s input, concentrating on the most influential interactions while improving prediction accuracy. The ground-truth label of behavioral intention \(\mathrm{H}^{*}\) is derived from an auto-labeled data preprocessing process (see Appendix Algorithm \autoref{alg:IntentionLabel}).

\subsubsection{Multimodal Vectorized Occupancy Prediction}
Unlike conventional occupancy prediction methods that rely on computationally intensive rasterization of multi-view images, we introduce a novel vectorized occupancy prediction framework that integrates seamlessly with our vectorized scenario representation. For each map polyline \(l_i\), we predict multimodal occupancy probabilities corresponding to different future hypotheses of the target agent. Denoting \(C^{i-1}\) as the previous vectorized occupancy tokens, we apply an operation symmetric to the multimodal behavioral intention prediction:
\begingroup
\small
\begin{equation}
\begin{aligned}
\hat{\mathrm{O}}^{i} 
&= \text{Sigmoid} \Bigl(\text{MLP} \bigl(C^{i} \bigr) \Bigr) 
\;\in\;\mathbb{R}^{K \times N_l \times 1}, \\
C^{i} &= \text{MLP}(\text{MLP}(L_4 \oplus Q^{i}) + C^{i-1} )
\end{aligned}
\end{equation}
\endgroup

 This vectorized approach ensures both efficiency and scalability while maintaining alignment with the overall vectorized representation of the scene. Among the \(N_l\) polylines's multimodal occupancy probabilities $\varphi(\hat{\mathrm{O}^{i}})\in \mathbb{R}^{K\times N_l}$, we select the top-\(n\) with the highest predicted probabilities in each modality to form \(L_{5}\in \mathbb{R}^{n\times D}\), where \(n \ll N_{l}\). These top-ranked polylines serve as focused inputs for the subsequent trajectory decoder. The ground-truth occupancy label \(\mathrm{O}^{*}\) is also derived from an auto-labeled data preprocessing process (see Appendix Algorithm \autoref{alg:vectorized_occupancy}).

\subsubsection{Trajectory Decoder}  
We adopt a multi-layer MTR-style\cite{shi2023MTR} trajectory decoder. At each layer \(i\), self-attention is applied to the query content \(Q^i \in \mathbb{R}^{K \times D}\) across the \(K\) motion modes, enabling information exchange among different future modalities. Subsequently, for each modality, two cross-attention modules integrate features from the filtered agent tokens \(A_5\) and map tokens \(L_5\). Finally, the target agent feature (replicated \(K\) times) is concatenated with the cross-attended query features, and passed through a regression head to generate a set of Gaussian Mixture Model (GMM) parameters at each future timestep: $
\Bigl\{ \bigl( \mu_x^k, \mu_y^k, \sigma_x^k, \sigma_y^k, \rho^k \bigr) \Bigr\}_{k=1}^K,$
where \(\bigl(\mu_x^k, \mu_y^k, \sigma_x^k, \sigma_y^k, \rho^k \bigr)\) parameterizes the \(k\)-th Gaussian component. In addition, a classification head outputs the confidence scores \(S \in \mathbb{R}^{K}\) corresponding to each motion mode. This multimodal representation captures the inherent uncertainties of agent trajectories. 

\subsection{Training Loss}
Our overall training objective comprises four components:
\begingroup
\small
\begin{equation}
\begin{aligned}
    \mathcal{L}_{\mathrm{total}} = \lambda_{\mathrm{1}}\,\mathcal{L}_{\mathrm{Int}}
    + \lambda_{\mathrm{2}}\,\mathcal{L}_{\mathrm{Occ}} + \lambda_{\mathrm{3}}\,\mathcal{L}_{\mathrm{Traj}}
    + \lambda_{\mathrm{4}}\,\mathcal{L}_{\mathrm{Score}},
\end{aligned}
\end{equation}
\endgroup
where \(\lambda_{\mathrm{1}}, \lambda_{\mathrm{2}}, \lambda_{\mathrm{3}}\), and \(\lambda_{\mathrm{4}}\) are weighting factors balancing the contributions of behavioral intention prediction, vectorized occupancy prediction, trajectory prediction, and mode classification, respectively. Specifically, \(\mathcal{L}_{\mathrm{Int}}\) is calculated using the multi-class Focal Loss, \(\mathcal{L}_{\mathrm{Occ}}\) is based on the binary Focal Loss, \(\mathcal{L}_{\mathrm{Traj}}\) is derived from the GMM loss, and \(\mathcal{L}_{\mathrm{Score}}\) is computed with Binary Cross-Entropy. During training, the winner-take-all strategy is applied for \(\mathcal{L}_{\mathrm{Int}}\), \(\mathcal{L}_{\mathrm{Occ}}\), and \(\mathcal{L}_{\mathrm{Traj}}\), ensuring that only the modality closest to the ground-truth trajectory is used to compute these losses. Please check appendix (\autoref{traininglossdetail}) for more details.\\

\section{Experiments}
\subsection{Experimental Setup}
\paragraph{Datasets and Evaluation Metrics.}
Our experiments are conducted on one of the most challenging prediction datasets, Waymo Open Motion Dataset (WOMD)\cite{waymo_motion_prediction}. This large-scale dataset comprises 486,995 training clips, 44,097 validation clips, and 44920 testing clips. Each clip contains 10 timesteps of historical agent states, 1 current timestep, and 80 future timesteps at a sampling frequency of 10 Hz, along with HD map information. We evaluate our method on both core WOMD tasks: the marginal motion and interactive motion prediction tasks. Following the standard evaluation protocol, we adopt metrics including Soft mAP, mAP, minADE, minFDE, Miss Rate, and Overlap Rate, and Soft mAP is the main metric.
\paragraph{Implementation Details.} We employ AdamW optimizer \cite{loshchilov2019decoupledweightdecayregularization} for training, conducting experiments on a cluster of 8 NVIDIA A800 GPUs with a total batch size of 80. The learning rate is initialized as $1\times10^{-4}$ and begins step decay starting at epoch 22, halving every two epochs. The model undergoes 30 epochs. More details can be referred to Appendix \autoref{network_detail}.

\subsection{Leaderboard Performance}
\begin{figure}[h]
    \centering
    \setlength{\abovecaptionskip}{-0.1cm}
        \includegraphics[width=0.8\linewidth]{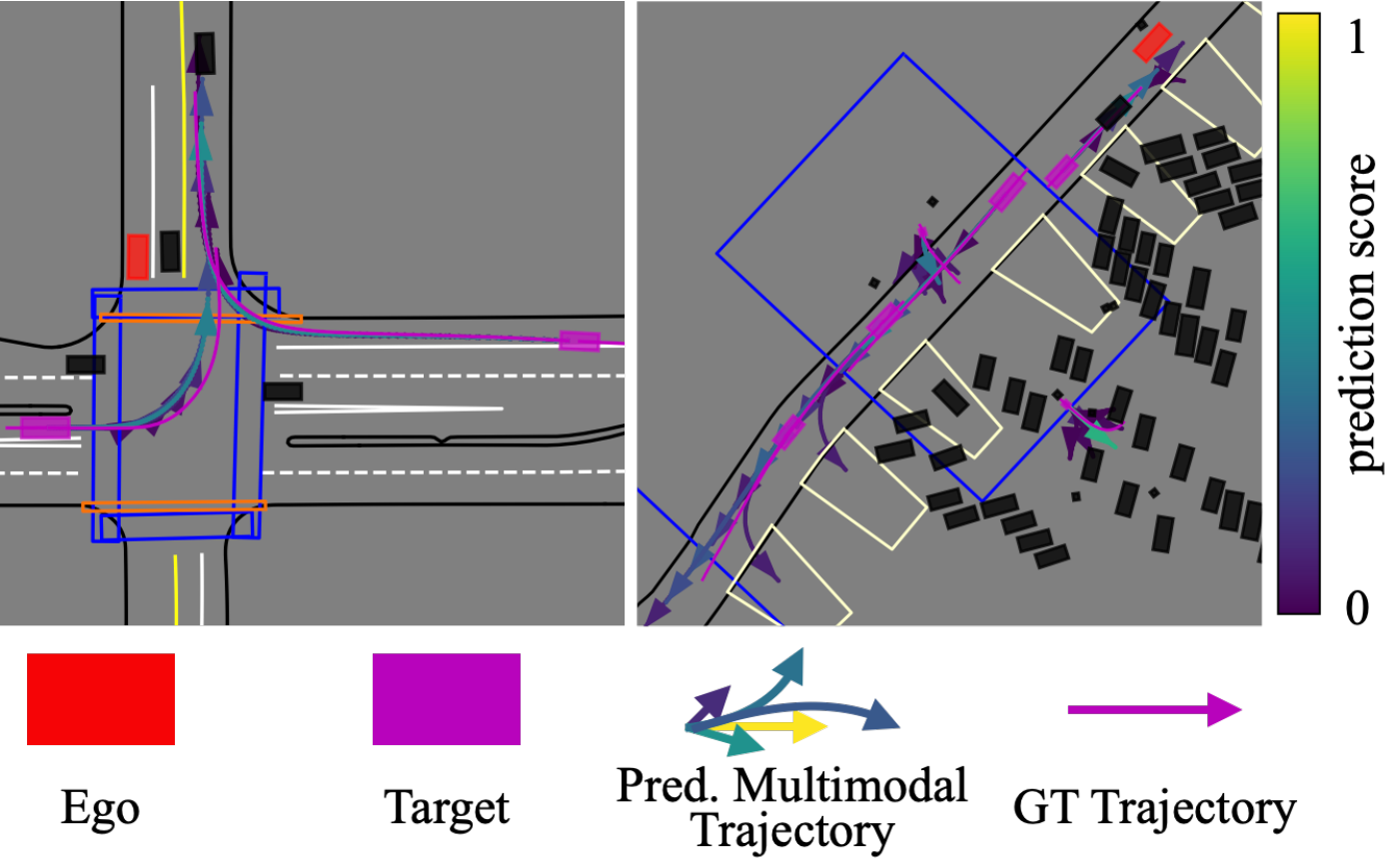}
    \hfill
    \caption{Visualization results for joint (left) and marginal (prediction) results. }
    \label{Majoint}
    \vspace{-8mm}
\end{figure}

\autoref{Majoint} illustrates the detailed qualitative results for both marginal and joint predictions using our proposed method. For additional qualitative results, please refer to \autoref{sec:Appendix Qualitative Results} in Appendix. 
\begin{table*}
\centering
\caption{Prediction on the test leaderboard of the motion prediction track of the Waymo Open Dataset Challenge. The first place is denoted by \textbf{bold}, the second place by \underline{underline}, and the third place by \textasteriskcentered{asterisk}.}
\label{tab:marginalleadboard}
\resizebox{0.7\linewidth}{!}{%
    \begin{tabular}{ccccccc}  
        \hline
        \rowcolor[rgb]{0.863,0.863,0.863} 
        Method &  Soft mAP $\uparrow$ & mAP $\uparrow$ & minADE $\downarrow$ & minFDE $\downarrow$ & Miss Rate $\downarrow$ & Overlap Rate $\downarrow$\\
        \hline
        \\[-1em]
        RMP-YOLO(Ensemble)\cite{sun2024rmpyolorobustmotionpredictor} & \underline{0.4737} & 0.4531 & \underline{0.5564} & \textbf{1.1188} & \textbf{0.1084} & ${0.1259}^{*}$ \\
        ModeSeq(Ensemble)\cite{zhou2024modeseq} & \underline{0.4737} & \textbf{0.4665} & 0.5680 & 1.1766 & 0.1204 & 0.1275 \\
        BeTop\cite{liu2024reasoningmultiagentbehavioraltopology} & 0.4698 & 0.4587 & 0.5716 & 1.1668 & 0.1183 & 0.1272 \\
        MGTR\cite{gan2024mgtrmultigranulartransformermotion} & 0.4599 & 0.4505 & 0.5918 & 1.2135 & 0.1298 & 0.1275 \\
        EDA\cite{lin2024eda} & 0.4596 & 0.4487 & 0.5718 & 1.1702 & 0.1169 & 0.1266 \\
        MTR++\cite{shi2024mtr++} & 0.4410 & 0.4329 & 0.5906 & 1.1939 & 0.1298 & 0.1281 \\
        
        MTR\cite{shi2023MTR} & 0.4403 & 0.4249 & 0.5964 & 1.2039 & 0.1312 & 0.1274 \\

        HPTR\cite{zhang2023hptr} & 0.3968 & 0.3904 & 0.5565 & 1.1393 & 0.1434 & 0.1366 \\
        HDGT\cite{JiaHDGT} & 0.3709 & 0.3577 & 0.5933 & 1.2055 & 0.1511 & 0.1557 \\
        DenseTNT\cite{gu2021densetnt} & - & 0.3281 & 1.0387 & 1.5514 & 0.1573 & 0.1779 \\
        SceneTransformer\cite{ngiam2022scenetransformer} & - & 0.2788 & 0.6117 & 1.2116 & 0.1564 & 0.1473 \\
        {\cellcolor[rgb]{0.902,0.902,0.902}}Ours (Ensemble) & {\cellcolor[rgb]{0.902,0.902,0.902}} \textbf{0.4801}
        & 
        {\cellcolor[rgb]{0.902,0.902,0.902}} ${0.4598}^{*}$ & {\cellcolor[rgb]{0.902,0.902,0.902}}\textbf{0.5563} & {\cellcolor[rgb]{0.902,0.902,0.902}}\underline{1.1295} & {\cellcolor[rgb]{0.902,0.902,0.902}}\underline{0.1087} & {\cellcolor[rgb]{0.902,0.902,0.902}}\underline{0.1258} \\
        {\cellcolor[rgb]{0.902,0.902,0.902}}Ours (Single) & {\cellcolor[rgb]{0.902,0.902,0.902}} 0.4721
        & 
        {\cellcolor[rgb]{0.902,0.902,0.902}}\underline{0.4609} & {\cellcolor[rgb]{0.902,0.902,0.902}}${0.5641}^{*}$ & {\cellcolor[rgb]{0.902,0.902,0.902}}${1.1540}^{*}$ & {\cellcolor[rgb]{0.902,0.902,0.902}}$ {0.1143}^{*}$ & {\cellcolor[rgb]{0.902,0.902,0.902}}\textbf{0.1255} \\
        \hline
        \\[-1em]
    \end{tabular}
}
\end{table*}

\begin{table*}
\centering
\caption{Joint Prediction on the test leaderboard of the interaction prediction track of the Waymo Open Dataset Challenge. The first place is denoted by \textbf{bold}, the second place by \underline{underline}, and the third place by \textasteriskcentered{asterisk}.}
\label{tab:jointleadboard}
\resizebox{0.7\linewidth}{!}{%
    \begin{tabular}{ccccccc}  
        \hline
        \rowcolor[rgb]{0.863,0.863,0.863} 
        Method & Soft mAP $\uparrow$ & mAP $\uparrow$ & minADE $\downarrow$ & minFDE $\downarrow$ & Miss Rate $\downarrow$ & Overlap Rate $\downarrow$ \\
        \hline
        \\[-1em]
        BeTop-ens \cite{liu2024reasoningmultiagentbehavioraltopology} & \underline{0.2573} & \underline{0.2511} & 0.9779 & 2.2805 & 0.4376 & $0.1688^{*}$ \\
        BeTop \cite{liu2024reasoningmultiagentbehavioraltopology} & $0.2466^{*}$ & $0.2412^{*}$ & 0.9744 & 2.2744 & 0.4355 & 0.1696 \\
        MTR++ \cite{shi2024mtr++} & 0.2368 & 0.2326 & $\underline{0.8975}$ & $\underline{1.9509}$ & \textbf{0.4143} & \textbf{0.1665} \\
        MTR \cite{shi2023MTR} & 0.2078 & 0.2037 & $0.9181^{*}$ & 2.0633 & 0.4411 & 0.1717 \\
        MotionDiffuser \cite{jiang2023motiondiffuser} & 0.2047 & 0.1952 & \textbf{0.8642} & \textbf{1.9482} & $\underline{0.4300}$ & 0.2004 \\
        GameFormer \cite{huang2023gameformergametheoreticmodelinglearning} & 0.1982 & 0.1923 & 0.9721 & 2.2146 & 0.4933 & 0.2022 \\
        BIFF \cite{zhu2023biffbilevelfuturefusion} & 0.1383 & 0.1229 & 0.9388 & $2.0443^{*}$ & 0.4620 & 0.2172 \\
        DenseTNT \cite{gu2021densetnt} & - & 0.1647 & 1.1417 & 2.4904 & 0.5350 & 0.2309 \\
        M2I \cite{sun2022m2ifactoredmarginaltrajectory} & - & 0.1239 & 1.3506 & 2.8325 & 0.5538 & 0.2757 \\
        SceneTransformer \cite{ngiam2022scenetransformer} & - & 0.1192 & 0.9774 & 2.1892 & 0.4942 & 0.2067 \\
        {\cellcolor[rgb]{0.902,0.902,0.902}}Ours (Single) & {\cellcolor[rgb]{0.902,0.902,0.902}}\textbf{0.2718} & {\cellcolor[rgb]{0.902,0.902,0.902}}\textbf{0.2659} & {\cellcolor[rgb]{0.902,0.902,0.902}}0.9738 & {\cellcolor[rgb]{0.902,0.902,0.902}}2.2734 & {\cellcolor[rgb]{0.902,0.902,0.902}}$0.4316^{*}$ & {\cellcolor[rgb]{0.902,0.902,0.902}}\underline{0.1684} \\
        \hline
    \end{tabular}
}
\end{table*}

\begin{figure*}
    \centering
    \setlength{\abovecaptionskip}{-0.2cm}
        \includegraphics[width=1\linewidth]{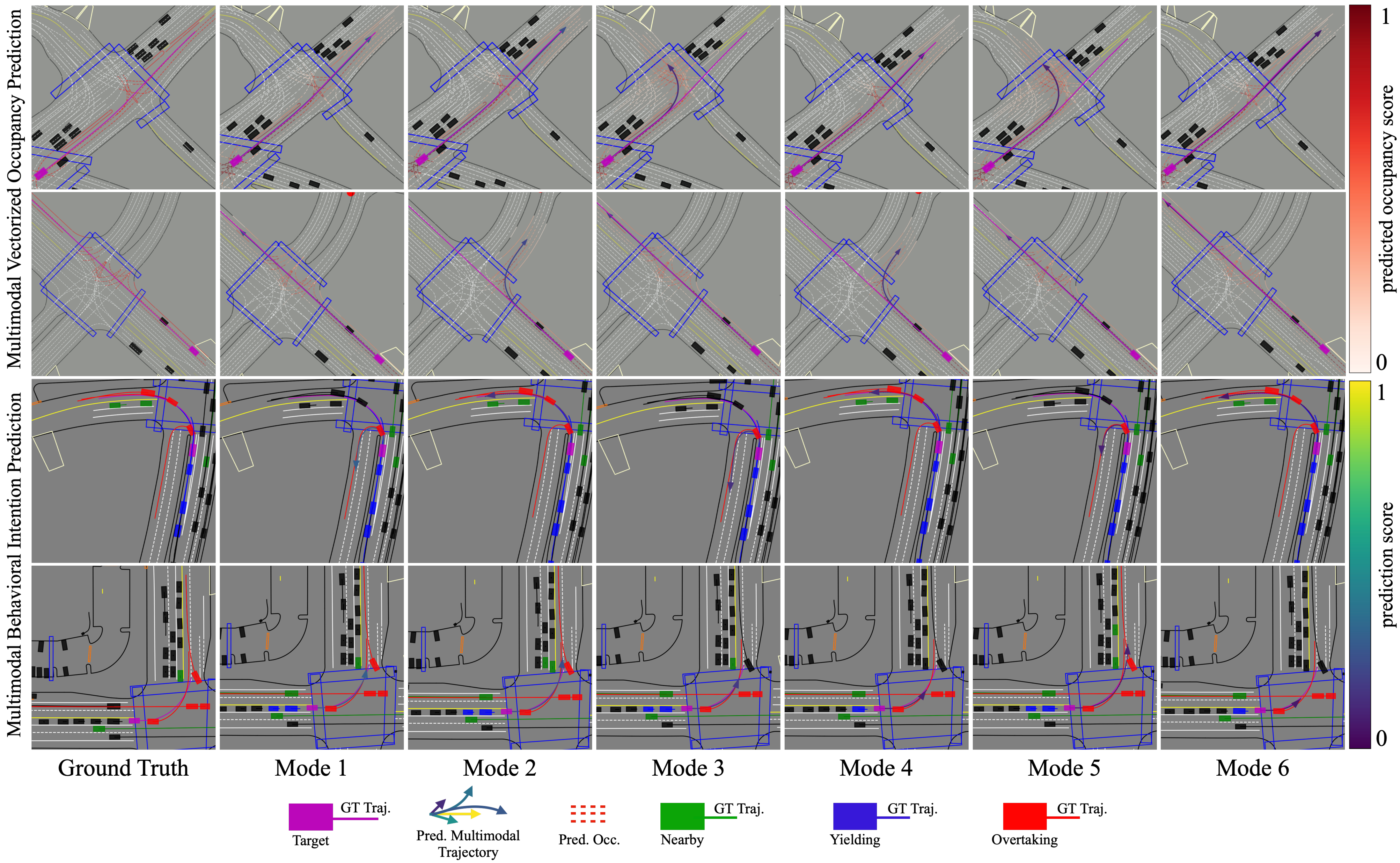}
    \hfill
\caption{Visualization of Predicted Multimodal Occupancy and Intention Labels.  
In the top two rows, black agents represent other agents, while in the bottom two rows, they indicate ignored agents. Ground-truth trajectories are included for validation of predicted behaviors.}
    \label{figure-main-bpop}
    \vspace{-4mm}
\end{figure*}
\subsection{Ablation Study}
\paragraph{Different behavior modeling and map interaction mechanisms.} 
\autoref{ablationstudy_selection} presents an ablation study evaluating various agent-pruning and map-pruning strategies on the validation set. The first row (All + Dynamic) queries all agents for the decoder while employing dynamic map selection as in MTR \cite{shi2023MTR}, serving as a baseline. Introducing agent selection via Braid Theory \cite{liu2024reasoningmultiagentbehavioraltopology} (Row 2) enhances Soft mAP to 0.4602. Replacing Braid Theory with our two-class filtering—where Yielding, Overtaking, and Nearby behaviors are treated as interactive while Ignore is considered non-interactive (Row 3)—further refines minADE to 0.5693 while maintaining a comparable Miss Rate. Notably, adopting our four-class YOIN approach (Row 4) elevates Soft mAP to 0.4728, demonstrating its ability to better capture nuanced interactions. Lastly, incorporating vectorized occupancy into YOIN (Row 5) further improves performance (Soft mAP = 0.4758), highlighting the benefit of explicit map filtering in isolating relevant polylines and refining trajectory predictions. These results validate the effectiveness of our integrated behavior modeling and map selection framework.

\begin{table*}
\centering
\caption{Ablation study of different agent-pruning and map-pruning strategies on the validation set, using the same proposed encoder.}
\label{ablationstudy_selection}

\resizebox{0.95\linewidth}{!}{
    \begin{tabular}{cccccccccccc}  
        \hline
        \rowcolor[rgb]{0.863,0.863,0.863} \multicolumn{4}{c}{Agent} & \multicolumn{2}{c}{Map} & \multicolumn{6}{c}{Metrics}  \\ 
        \rowcolor[rgb]{0.863,0.863,0.863} All & w/Braid Theory\cite{liu2024reasoningmultiagentbehavioraltopology} & w/Inter. & w/YOIN & w/Dynamic\cite{shi2023MTR} & w/Vect. Occ. & Soft mAP & mAP & minADE & Miss Rate & Inference Time (ms/scenario) & Params (M) \\ \hline 
        \checkmark & ~ & ~ & ~ & \checkmark & ~ & 0.4582 & 0.4414 & 0.5718 & 0.1203 & 14.67 & 69.469 \\ 
        ~ & \checkmark & ~ & ~ & \checkmark & ~ & 0.4602 & 0.4507 & 0.5745 & 0.1187 & 23.92 & 48.938 \\ 
        ~ & ~ & \checkmark & ~ & \checkmark & ~ & 0.4663 & 0.4549 & 0.5693 & 0.1187 & 23.92 & 48.938 \\ 
        ~ & ~ & ~ & \checkmark & \checkmark & ~ & 0.4728 & 0.4613 & 0.5680 & 0.1163 & 23.92 & 48.941 \\ 
        ~ & ~ & ~ & \checkmark & ~ & \checkmark & 0.4758 & 0.4646 & 0.5652 & 0.1173 & 18.27 & 48.944 \\ 
        \hline
    \end{tabular}
}
\end{table*}
\paragraph{Marginal Prediction Performance.} \autoref{tab:marginalleadboard} shows that our method achieves state-of-the-art (SOTA) performance on the Waymo 2024 motion prediction benchmark, outperforming all existing LiDAR-free approaches in the primary metric, Soft mAP, as well as in minADE and Overlap Rate. In the remaining metrics (mAP, minFDE, and Miss Rate), our method ranks second. Here, 'Ensemble' denotes the model-ensemble technique described in \cite{shi2023MTR} for performance boosting, while 'Single' indicates evaluation with a single model.
\paragraph{Joint Prediction Performance.} 
As presented in \autoref{tab:jointleadboard}, even without any model-ensemble techniques \cite{shi2023MTR}, our single model achieves the best performance on the Waymo joint prediction leaderboard, surpassing the previous SOTA method \cite{liu2024reasoningmultiagentbehavioraltopology} by 10.2\% in both Soft mAP and mAP. Furthermore, our model attains the second-lowest Overlap Rate and the third-lowest Miss Rate. These substantial improvements underscore the effectiveness of the IMPACT framework.

\begin{table}
\centering
\caption{Multimodal Behavioral Intention Prediction Performance.}
\label{tab:YOI prediction}
\resizebox{0.99\linewidth}{!}{%
    \begin{tabular}{cccccc} 
        \hline
        \rowcolor[rgb]{0.863,0.863,0.863} {\cellcolor[rgb]{0.863,0.863,0.863}} & \multicolumn{3}{c}{Top-1} & Top-6 & {\cellcolor[rgb]{0.863,0.863,0.863}} \\
        \rowcolor[rgb]{0.863,0.863,0.863} \multirow{-2}{*}{{\cellcolor[rgb]{0.863,0.863,0.863}}Class} & Precision & Recall & F1-Score & Accuracy & \multirow{-2}{*}{{\cellcolor[rgb]{0.863,0.863,0.863}}GT Data Ratio (\%)} \\ 
        \hline
        Ignored & 0.99 & 0.97 & 0.98 & 0.99 & 89.12 \\
        Nearby & 0.71 & 0.89 & 0.79 & 0.97 & 3.93 \\
        Overtaking & 0.82 & 0.89 & 0.85 & 0.97 & 2.59 \\
        Yielding & 0.86 & 0.96 & 0.90 & 0.97 & 4.32 \\ 
        \hline
        All & 0.97 & 0.97 & 0.97 & 0.987& 100.00 \\
        \hline
    \end{tabular}
}
\end{table}
\subsection{Behavioral Intention Prediction Performance}

\autoref{tab:YOI prediction} presents the results of multimodal behavioral intention prediction across four designated categories. Overall, the model demonstrates strong performance, with a F1-Score of 0.97, and a Top-6 accuracy of 0.987. In the dominant \textit{Ignored} category, which comprises 89.12\% of the dataset, our method attains particularly high accuracy (F1 = 0.98). Despite being underrepresented, the \textit{Nearby}, \textit{Overtaking}, and \textit{Yielding} classes achieve F1-Scores between 0.79 and 0.90, showcasing the model’s robustness in handling less frequent behaviors. These results underscore the model’s effectiveness in accurately identifying and predicting diverse driver intentions, offering valuable insights for interpretable prediction models and informed downstream decision-making.

\subsection{Vectorized Occupancy Prediction Performance}
\begin{table}[H]
\centering
\caption{Binary Classification Performance for top-1 mode Occupancy Prediction.}
\label{tab:occ_prediction}
\resizebox{0.9\linewidth}{!}{%
    \begin{tabular}{ccccc}
        \hline
        \rowcolor[rgb]{0.863,0.863,0.863} Class & Precision & Recall & F1-Score & GT Data Ratio (\%) \\ \hline
        Occupied & 0.933 & 0.775 & 0.847 & 3.96 \\
        Unoccupied & 0.991 & 0.998 & 0.994 & 96.04 \\
        \hline
        All & \multicolumn{3}{c}{0.989} & 100.00 \\
        \hline
    \end{tabular}
}
\vspace{-3mm}
\end{table}
\autoref{tab:occ_prediction} presents precision, recall, and F1-score for both classes, along with their data ratios in the validation set. The \emph{occupied} class constitutes only 3.96\% of samples, highlighting a strong imbalance. Despite this, the model achieves a high overall accuracy of 0.989. 
For the \emph{occupied} class, precision (0.933) and recall (0.775) yield an F1-score of 0.847, indicating rare false positives—crucial for pruning irrelevant map polylines in trajectory prediction. The dominant \emph{unoccupied} class achieves near-perfect precision (0.991), recall (0.998), and F1-score (0.994). These results demonstrate that our vectorized occupancy prediction effectively identifies critical map segments while minimizing false classification. More qualitative results (like \autoref{figure-main-bpop}) are in Appendix \autoref{sec:Appendix Qualitative Results}.

\begin{figure}
    \centering
\includegraphics[width=1\linewidth]{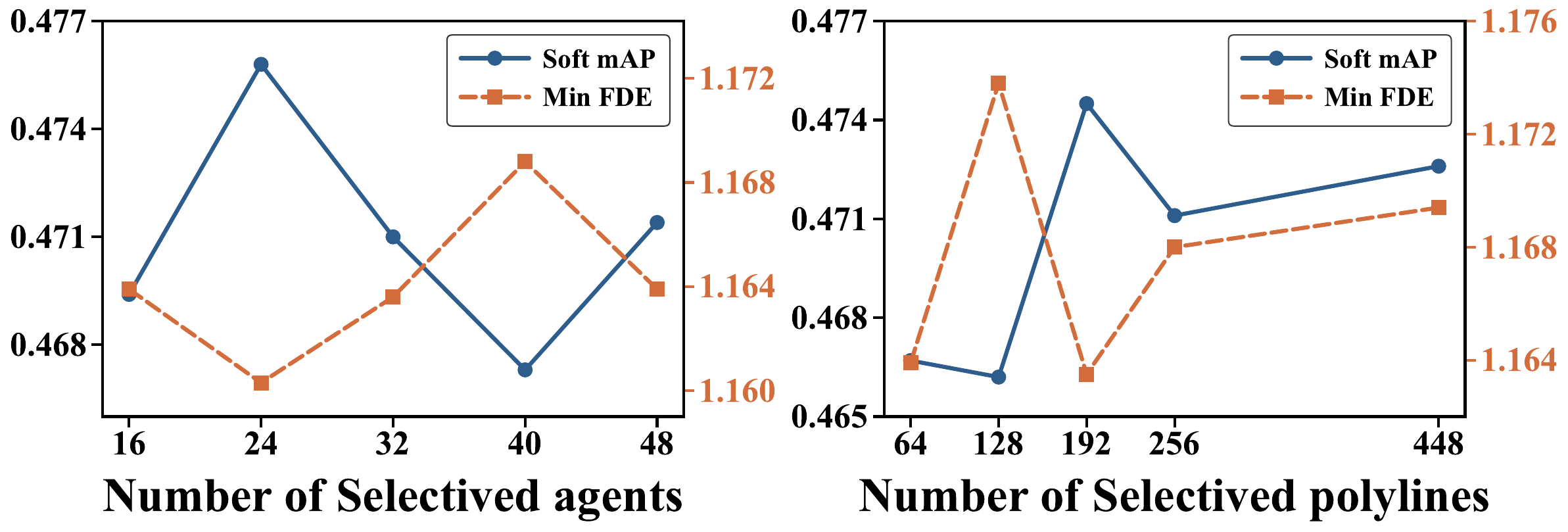}
    \caption{Results of varying numbers of selected agents and polylines for modeling on validation dataset.}
    \label{figure-sampling-space}
    \vspace{-6mm}
\end{figure}

\paragraph{Number of selected agents and map polylines for behaviour-guided and occupancy-guided decoder.} As shown in \autoref{figure-sampling-space}, selecting 24 agents and 192 polylines achieves the optimal balance between interaction diversity and trajectory accuracy. On average, the model input contains 43.85 agents and 749.52 map polylines, meaning this selection reduces scene complexity by 45.2\% and 74\%, respectively, while preserving essential contextual cues. Beyond this threshold, redundant elements introduce noise, degrading generalization and increasing prediction uncertainty. This highlights the importance of selective context pruning to enhance trajectory forecasting performance.

\paragraph{Computational Efficiency and Parameter Analysis.}
We compare various agent-pruning and map-pruning strategies in \autoref{ablationstudy_selection} using the same encoder architecture. The first row (All + w/Dynamic) corresponds to a baseline that queries all agents and dynamically collect polylines, resulting in a model with 69.469 M parameters and a reported inference time of 14.67 ms under a simplified setting.
By contrast, our final approach (the last row in \autoref{ablationstudy_selection}) combines YOIN-based agent pruning with vectorized occupancy-based pruning. This design lowers the parameter count to 48.944 M while reducing the average inference time of BeTOP\cite{liu2024reasoningmultiagentbehavioraltopology} from 23.92 to 18.27 ms per scenario. The key to these efficiency gains lies in pruning irrelevant scene information—both agents and polylines—based on their semantic contributions, which in turn eliminates redundant computations.

\begin{table}[H]
\centering
\caption{Performance gains from integrating proposed modules with existing approaches(Argoverse1\&2). bFDE6 is main metric.}
\label{tab:Generalization}
\resizebox{0.99\linewidth}{!}{
    \begin{tabular}{ccccc} 
        \hline
        \rowcolor[rgb]{0.863,0.863,0.863} Method & bFDE\textsubscript6 & mADE\textsubscript6 & mFDE\textsubscript6 & MR \\ 
        \hline
        SIMPL (AV2) & 2.069 & 0.777 & 1.452 & 0.196 \\ 
        SIMPL+Ours (AV2) & 1.921 (-7.1\%) & 0.743 & 1.387 & 0.178 \\
        \hline
        HiVT (AV1) & 1.662 & 0.661 & 0.969 & 0.092 \\ 
        HiVT+Ours (AV1) & 1.556 (-6.4\%) & 0.599 & 0.932 & 0.087 \\
        \hline
    \end{tabular}%
    }
\vspace{-3mm}
\end{table}

\textbf{Generalization Ability Study}. To validate the cross-dataset and cross-paradigm generalization of our proposed symmetric dual filter, we conduct transfer experiments between the Waymo Open Dataset and Argoverse (AV1/AV2) using two distinct baselines: SIMPL (query-centric) and HiVT (agent-centric). As shown in \autoref{tab:Generalization}, our approach achieves up to 7.1\% (SIMPL on AV2) and 6.4\% (HiVT on AV1) improvements in bFDE\textsubscript{6}. These results demonstrate our method’s versatility as a general enhancement framework, independent of specific network architectures.

\section{Conclusion}
In this paper, we present IMPACT, a novel and unified module that advances multimodal motion prediction through explicit modeling of behavioral intentions and dynamic context optimization. The joint intention and motion modeling module eliminates redundancy and enables seamless information flow between behavioral semantic and motion dynamics, the experiments on both marginal and joint motion prediction challenges of large-scale WOMD show that our approach achieves state-of-the-art performance. The adaptive pruning decoder leverages intention and occupancy prediction priors to reduce computational complexity while preserving essential interaction cues. The automated labeling framework generates intention annotations across mainstream datasets and shows great scalable performance.\\
\textbf{Limitations and Future work.} Our pruning mechanism prioritizes instantaneous interactions, potentially overlooking evolving multi-agent gaming dynamics. Extending the framework with temporal graph networks or recursive reasoning could enhance long-horizon interaction modeling. We plan to incorporate this idea into both the latest end-to-end and traditional planning frameworks, particularly to address the challenge of determining which agent to predict and which polyline to focus on. 
\bibliographystyle{ieeenat_fullname}
\bibliography{main}

\clearpage
\setcounter{page}{1}
\maketitlesupplementary
\setcounter{section}{0}  

\section{Behavioral Intention Categories Split.}
\label{CategoriesReasons}
The selection of four intention labels: nearby, ignore, overtaking, and yielding—is motivated by balancing model expressiveness, computational efficiency, and actionable interpretability for downstream decision-making.\\
\begin{enumerate}
    \item \textbf{Limitations of Oversimplified Labels (Interactive/Ignore)}. While a binary classification reduces complexity, it fails to capture nuanced interaction types, agents labeled as "interactive" may exhibit vastly different kinematic patterns (e.g., merging vs. lane-keeping), which cannot be resolved without finer-grained labels.
    \item \textbf{Drawbacks of Overly Complex Labels}. Complex taxonomies struggle to generalize across diverse interaction patterns. Also, high-dimensional label spaces dilute feature representations, reducing inter-class separability. Lastly, rare categories lack sufficient training samples, causing models to memorize noise rather than learn generalizable patterns.
    \item \textbf{The rationality of the four categories}. According to algorithm\ref{alg:IntentionLabel}, these four labels have corresponding interaction information, which covers most interaction modes in urban driving. 4D embeddings also achieve optimal trade-offs between class separation and parameter efficiency.
\end{enumerate}

\section{Training Loss Detail}
\label{traininglossdetail}
The overall training objective comprises four components: $\mathcal{L}_{\mathrm{Int}}, \mathcal{L}_{\mathrm{Occ}},\mathcal{L}_{\mathrm{Traj}}$ and $ \mathcal{L}_{\mathrm{Score}}$. Since $\mathcal{L}_{\mathrm{Int}}, \mathcal{L}_{\mathrm{Occ}},\mathcal{L}_{\mathrm{Traj}}$ are multimodal to avoid mode-collapse, we apply winner-takes-all strategy. The weighting factors are set as \(\lambda_{\mathrm{1}}=100\), \(\lambda_{\mathrm{2}}=100\), \(\lambda_{\mathrm{3}}=1\), and \(\lambda_{\mathrm{4}}=1\).
\subsection{Behavioral Intention Prediction.}
The behavioral intention module predicts \(\hat{H}^{(k)} \in \mathbb{R}^{N_a \times 4}\) for each modality \(k\) and each agent, and the four output classes represent different intentions (yielding, overtaking, ignore, and nearby). The ground-truth label \(H^* \in \mathbb{R}^{N_a \times 4}\) is a one-hot encoding of the true class for each agent. Once the winning modality \(k^*\) is determined, we use \(\hat{H}^{(k^*)}\) to compute the Focal Loss \cite{lin2018focallossdenseobject} in a multi-class setting:
\begin{equation}
\mathcal{L}_{\mathrm{int}} 
= 
-\sum_{n=1}^{N_a} 
\sum_{c=1}^{4}
\alpha_c 
\Bigl(1 - \hat{H}_{n,c}^{(k^*)}\Bigr)^{\gamma_c}
\,H_{n,c}^* \,
\log\bigl(\hat{H}_{n,c}^{(k^*)}\bigr),
\end{equation}

where \(\alpha_c= [0.1, 0.45, 0.45, 0.45]\) and \(\gamma_c= [2, 1, 1, 1]\) are class-dependent balancing and focusing parameters, respectively. Intuitively, \(\alpha_c\) compensates for class imbalance (i.e., classes with fewer training examples receive higher weight), while \(\gamma_c\) “focuses” more strongly on hard, misclassified examples.

\subsection{Vectorized Occupancy Prediction.}

In the vectorized occupancy module, the prediction \(\hat{O}\in\mathbb{R}^{N_l\times1}\) represents the probability that each of the \(N_l\) map polylines is occupied by the target agent in the future. The ground-truth label is \(O^*\in \{0,1\}^{N_l\times1}\). We again use only the winning modality’s output \(\hat{O}^{(k^*)}\) and compute a binary Focal Loss:

\begin{equation}
\begin{aligned}
\mathcal{L}_{\mathrm{occ}} 
= 
&- \alpha 
\sum_{l=1}^{N_l}
\Bigl[
  O_l^*
  \Bigl(1 - \hat{O}_l^{(k^*)}\Bigr)^\gamma 
  \log\Bigl(\hat{O}_l^{(k^*)}\Bigr)\\
  &+
  \bigl(1 - O_l^*\bigr)
  \Bigl(\hat{O}_l^{(k^*)}\Bigr)^\gamma
  \log\bigl(1-\hat{O}_l^{(k^*)}\bigr)
\Bigr].
\end{aligned}
\end{equation}
where \(\alpha=0.25\) and \(\gamma=2\). 
This choice of Focal Loss mitigates the class imbalance between “occupied” and “not occupied” labels and focuses the training on hard-to-predict polylines.

\subsection{GMM-based Trajectory Loss.}
Given the predicted future Gaussian components \(\hat{\mathrm{Y}} \in \mathbb{R}^{K \times T_f \times 5}\), each modality \(k\) predicts parameters \(\bigl(\mu_x^{(k)}, \mu_y^{(k)}, \sigma_x^{(k)}, \sigma_y^{(k)}, \rho^{(k)}\bigr)\) for a bivariate Gaussian distribution at each future timestep \(t\). Let the ground-truth future trajectories be \(\mathrm{Y}^{*} = \{(x_t^*, y_t^*)\}_{t=1}^{T_f} \in \mathbb{R}^{T_f \times 2}\). We compute the Negative Log-Likelihood (NLL) loss only for the selected (winning) modality \(k^*\):

\begin{equation}
\mathcal{L}_{\mathrm{traj}}
= 
-\sum_{t=1}^{T_f}
\log\Bigl(
  \mathcal{N}\bigl(\mathbf{x}_t^*;\,
    \boldsymbol{\mu}_t^{(k^*)}, 
    \boldsymbol{\Sigma}_t^{(k^*)}
  \bigr)
\Bigr),
\end{equation}

where \(\boldsymbol{\mu}_t^{(k^*)} = [\mu_{x,t}^{(k^*)}, \mu_{y,t}^{(k^*)}]\) and covariance matrix \(\boldsymbol{\Sigma}_t^{(k^*)}\) is defined as:

\begin{equation}
\boldsymbol{\Sigma}_t^{(k^*)} = 
\begin{bmatrix}
\sigma_{x,t}^{(k^*)2} & \rho_t^{(k^*)}\sigma_{x,t}^{(k^*)}\sigma_{y,t}^{(k^*)} \\
\rho_t^{(k^*)}\sigma_{x,t}^{(k^*)}\sigma_{y,t}^{(k^*)} & \sigma_{y,t}^{(k^*)2}
\end{bmatrix}.
\end{equation}

\subsection{Mode Classification Loss.}
Additionally, each modality \(k\) predicts a confidence score \(\hat{S_k}\in [0,1]\). We wish to train these scores so that the winning modality receives higher scores while others receive lower scores. Let \(\delta^*\in \{0,1\}^K\) be a label vector that has \(\delta_{k^*}^*=1\) for the winning modality and \(\delta_{k}^*=0\) for all others. We use the standard binary cross-entropy:

\begin{equation}
\mathcal{L}_{\mathrm{score}} 
=
- \sum_{k=1}^K 
\Bigl[
  \delta_k^*
  \log(\hat{S_k})
  +
  (1 - \delta_k^*)
  \log\bigl(1 - \hat{S_k}\bigr)
\Bigr].
\end{equation}

\begin{figure*}
    \centering
    \includegraphics[width=16cm]{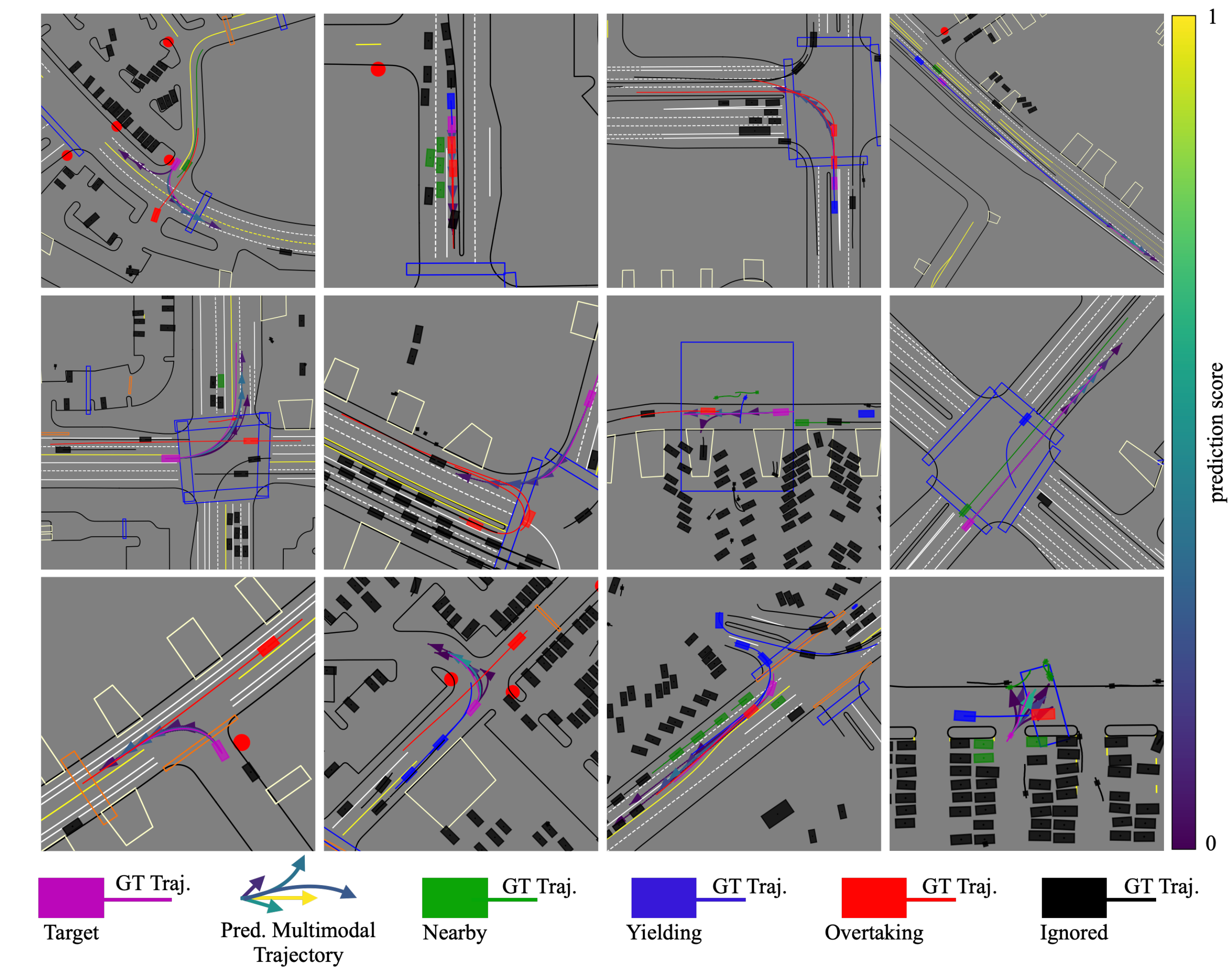} 
    \caption{Visualization of multimodal prediction results for the target agent, along with the top-1 mode behavioral intention predictions of surrounding agents. Ground-truth trajectories are also visualized to facilitate clear validation of the predicted behaviors.}
    \label{fig:yoi}
    \vspace{-1em}
\end{figure*}

\begin{figure*}
    \centering
    \includegraphics[width=16cm]{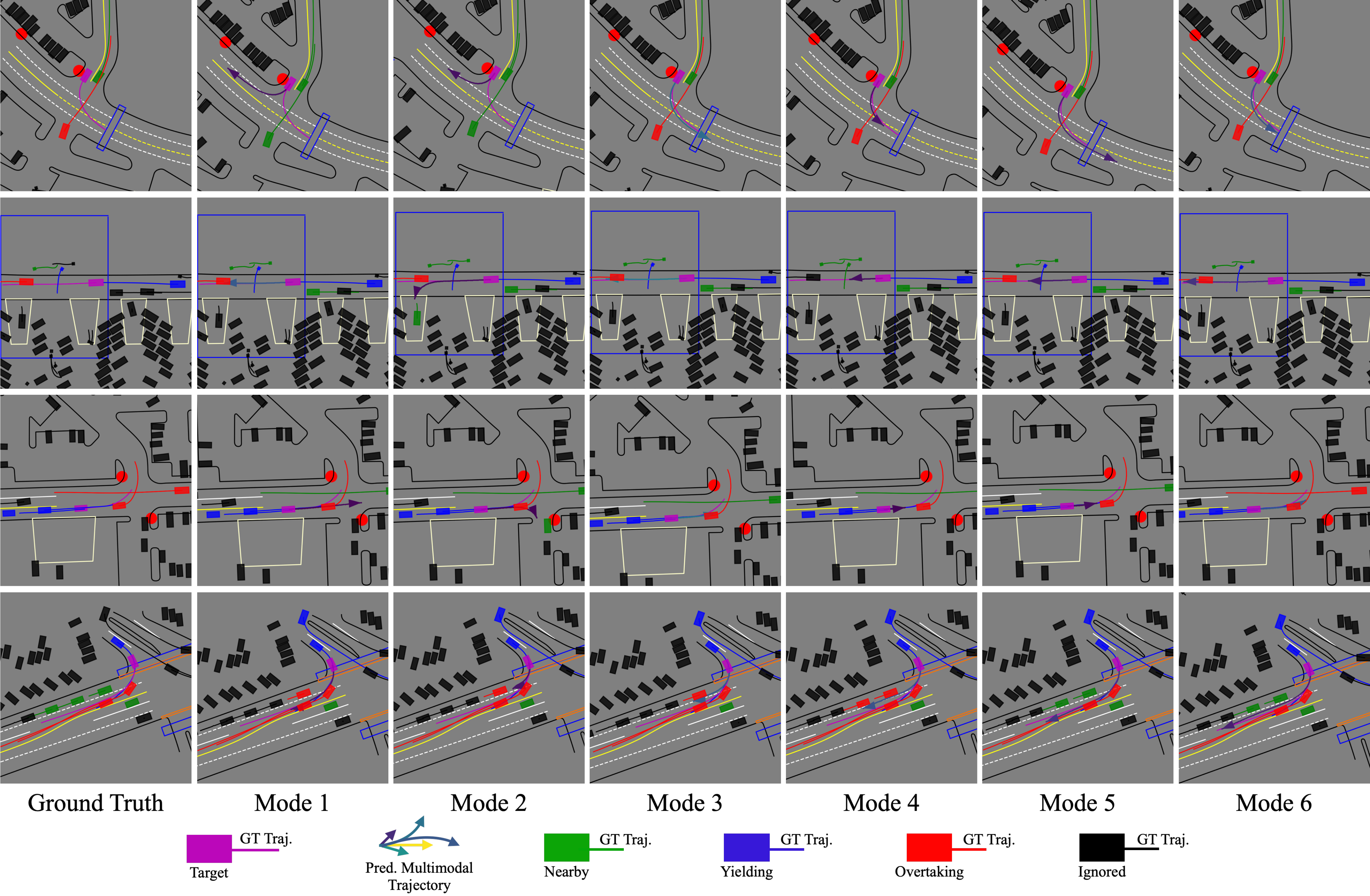} 
    \caption{Visualization of predicted multi mode intention labels. The first column renders the ground truth intention labels of the agents. Remaining columns render the predicted results of K=6 different modes. We can see that our predicted intentions and trajectories are coupled and cover most possible modes.}
    \label{fig:mode}
    \vspace{-1em}
\end{figure*}

\begin{figure*}
    \centering
    \includegraphics[width=16cm]{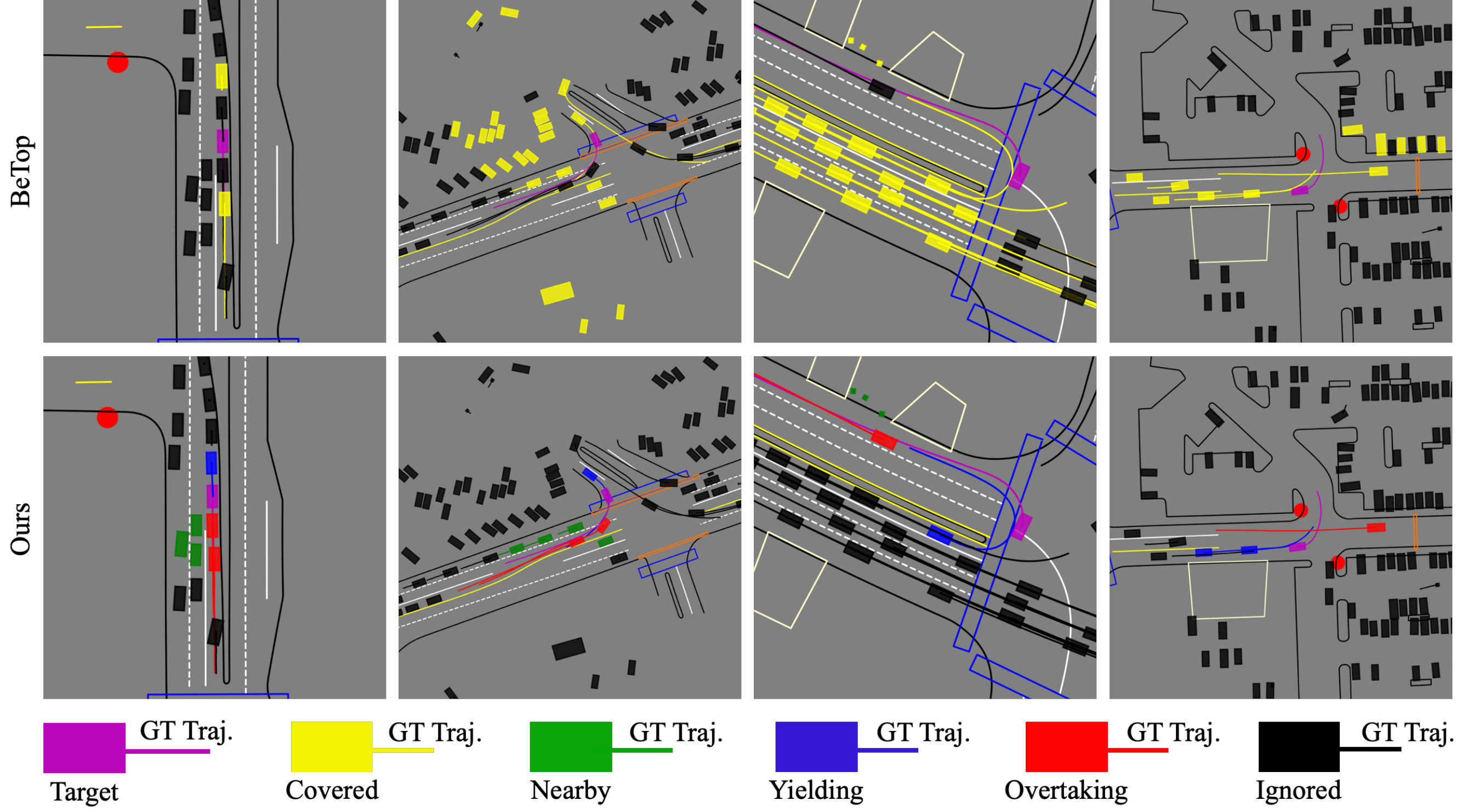} 
    \caption{Visualization of ground truth behaviour intention labels comparison between BeTop(Covered, Ignored) and IMPACT(Nearby, Yielding, Overtaking, Ignored). The first row of the figures renders the result of BeTop.}
    \label{fig:comparebetop}
    \vspace{-1em}
\end{figure*}

\begin{figure*}
    \centering
    \includegraphics[width=16cm]{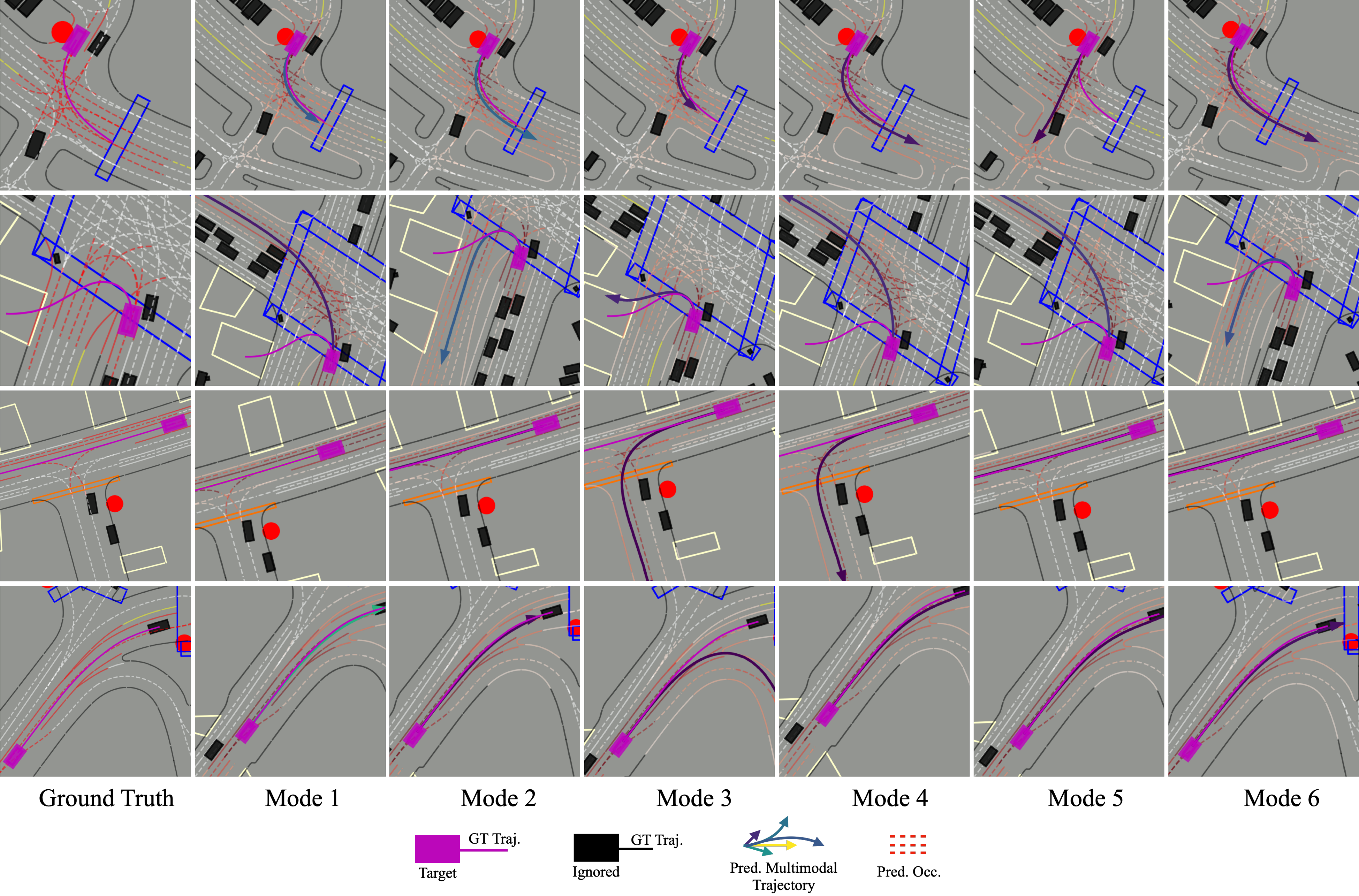} 
    \caption{Visualization of predicted multi modes vectorized occupancy. The first column renders the ground truth vectorized occupancy along the agents' future trajectories. Remaining columns render the predicted results of K=6 different modes.}
    \label{fig:compocc}
    \vspace{-1em}
\end{figure*}

\begin{figure*}
    \centering
    \includegraphics[width=16cm]{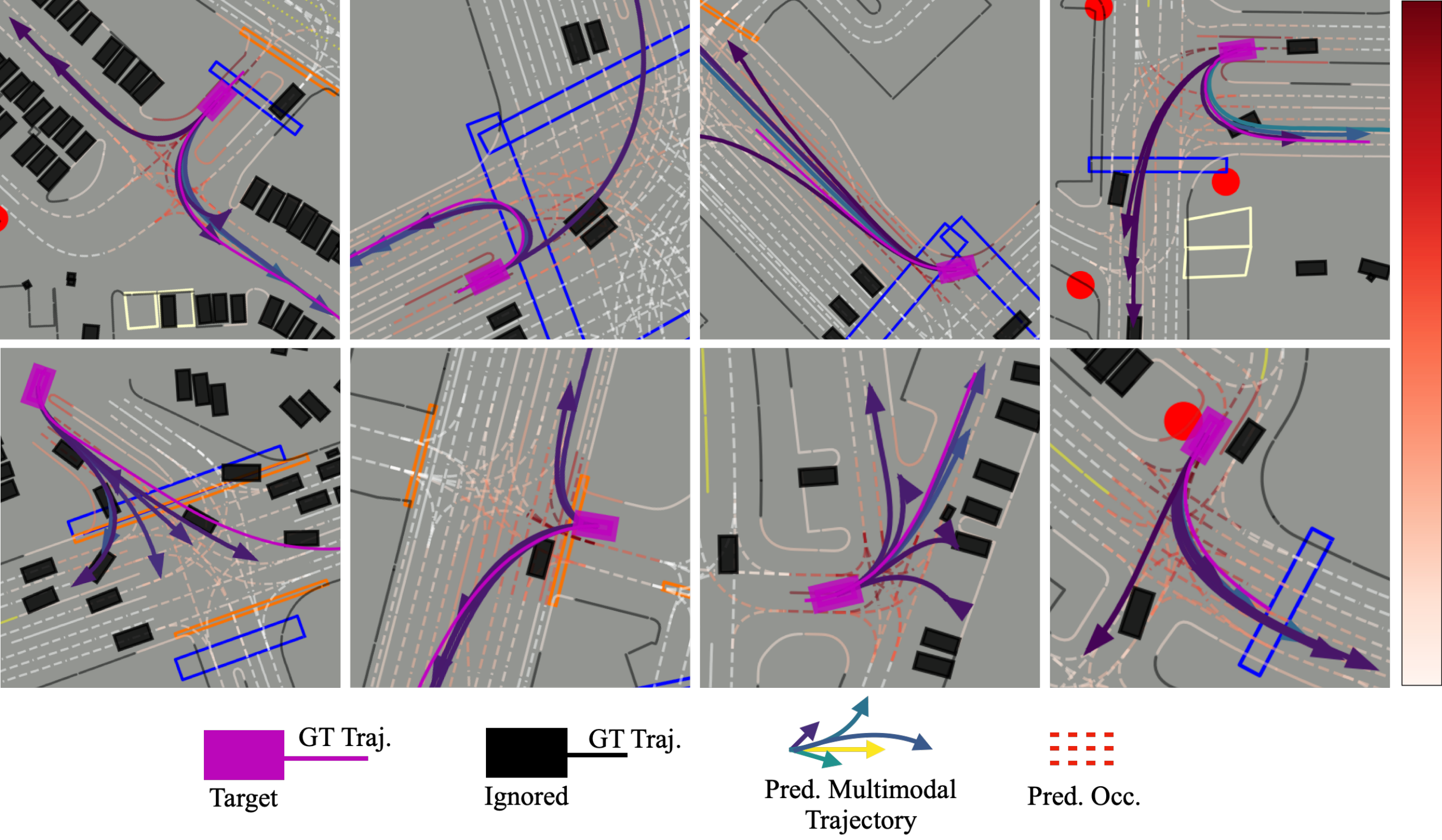} 
    \caption{Visualization of multimodal predicted vectorized occupancy for the target agents in different scenarios. The darker the polylines, the greater the probability of occupation.}
    \label{fig:predocc}
    \vspace{-1em}
\end{figure*}

\section{Model Ensemble Details.}
Given \(N\) well-trained models, each model produces 64 distinct future trajectories, resulting in a total of \(64N\) multimodal trajectories for each target agent. Each trajectory is assigned a confidence score by its originating model.

First, we apply a softmax normalization to the confidence scores of all \(64N\) trajectories. Next, we perform non-maximum suppression (NMS) based on the endpoints of these trajectories to select the top six. The NMS distance threshold \(\sigma\) scales according to the length \(L\) of the highest-confidence trajectory among the \(64N\) predictions, as follows:

\begin{equation}
\sigma = \min \left(3.5, \max \left(2.5, \frac{L - 10}{50 - 10} \times 1.5 + 2.5 \right) \right).
\end{equation}

 The performance of IMPACT(Ensemble) is generated by an ensemble of four different models:
\begin{enumerate}
    \item IMPACT model.
    \item IMPACT without Behavioral Intention Prediction.
    \item IMPACT without IMPACT Occupancy Prediction.
    \item IMPACT with QCNet refinement module.
\end{enumerate}

\section{Network Details}
\label{network_detail}
The agent motion encoder utilizes a Multi-Scale LSTM (MSL) module with three parallel Conv1D-LSTM streams (kernel sizes $\{1,3,5\}$) to process 30-dimensional historical states. Each Conv1D layer extracts 64-channel features, which are fed into LSTMs (64$\rightarrow$256D) and subsequently concatenated across scales before being projected to 256D motion tokens via an MLP. The relative motion encoder follows an identical structure, processing 4D inputs to generate 256D output tokens. Map polylines, represented by 9D features, are processed through a 5-layer MLP (64 hidden units per layer) with max-pooling and linear projection to 256D. Two MCG layers then fuse the agent, map, and relative motion features, preserving the 256D input-output dimensionality. The transformer encoder consists of 6 layers, each employing 8 attention heads with a dropout rate of 0.1 and K nearest neighbor attention (K=16). The behavioral intention module processes the encoded features through 256$\rightarrow$128D hidden layers to produce 4-class probability distributions ($\mathbf{h}_i \in \mathbb{R}^4$) for the categories \textit{ignore/nearby/overtaking/yielding}. The utility function \(\psi\)$(\cdot)$ applies weighted sum of softmax normalization to rank agents, assigning weights [0,1,1,1] to the respective categories, and selects the top-$m$ ($m=24$) agents with the highest scores for further processing. In parallel, the vectorized occupancy module uses 256$\rightarrow$128D hidden layers to predict the occupancy probabilities of map polylines. The utility function \(\varphi\)$(\cdot)$ applies sigmoid activation to these probabilities and retains the top-$n$ ($n=192$) polylines with the highest likelihood of occupancy. Both the selected agents and map elements (256D features) are then passed to the trajectory decoder, which comprises 6 iterative layers each equipped with 8 attention heads and 512-dimensional hidden states. Within each decoder layer, the multi-head attention mechanism operates with the same 0.1 dropout rate as the encoder, enabling context-aware feature refinement. Every layer independently predicts intermediate trajectory parameters ($\mu_x$, $\mu_y$, $\sigma_x$, $\sigma_y$, $\rho$) and confidence scores through dedicated three-layer MLP heads (512 hidden units per layer).

\section{Autolabel Algorithms for Behavioral Intention and Vectorized Occupancy}
\label{sec:rationale}
The autolabeling algorithms consist of two parts, Vectorized GT Occupancy Label Generation \ref{alg:vectorized_occupancy} and GT behavioral Intention Label Generation \ref{alg:IntentionLabel}.
\begin{algorithm}
\caption{Vectorized GT Occupancy Label Generation}
\label{alg:vectorized_occupancy}
\begin{algorithmic}[1]
    \REQUIRE ~~\\
    \quad Map polyline points $L \in \mathbb{R}^{N_l \times N_p \times 2}$; \\
    \quad Target agent's future trajectory $y \in \mathbb{R}^{T_f \times 2}$; \\
    \quad Distance threshold $\alpha$; \\
    \ENSURE ~~\\
    \quad Occupancy labels $O \in \{0,1\}^{N_l \times 1}$;
    \STATE Broadcast $L$ to $[\,N_l,\,N_p,\,1,\,2\,]$ and $y$ to $[\,1,\,1,\,T_f,\,2\,]$.
    \STATE Compute the distance tensor $d \in \mathbb{R}^{N_l \times N_p \times T_f}$.
    \STATE $O = (\min(d, \text{axis}=1) \leq \alpha\;)\text{.any}(\text{axis}=-1)$.
    \RETURN $O\in \mathbb{R}^{N_{l} \times 1}$
\end{algorithmic}
\end{algorithm}

\floatname{algorithm}{Algorithm}
\renewcommand{\algorithmicrequire}{\textbf{Input:}}
\renewcommand{\algorithmicensure}{\textbf{Output:}}
\begin{algorithm}
\caption{GT behavioral Intention Label Generation}
\label{alg:IntentionLabel}
\begin{algorithmic}[1]
    \REQUIRE ~~\\
        Agents' future trajectories $Y^{*} \in \mathbb{R}^{N_a \times T_f \times 2}$; \\
        Target agent's future trajectory $y \in \mathbb{R}^{T_f \times 2}$;
    \ENSURE ~~\\
        behavioral Intention labels $H \in \mathbb{R}^{N_a \times 4}$;
      
    \STATE \textbf{Radius Screening:}
    \FOR{$i = 1 \dots N_a$}
      \STATE Compute $d_i(t) = \| Y^{*}_{i,t} - y_t\|_2,\;\forall\,t = 1 \dots T_f.$
      \IF{$\min_{t} \, d_i(t) > \tau_{\mathrm{ignore}}$}
        \STATE Label $i$ as $\texttt{ignored}$ and exclude from further checks.
      \ELSE
        \STATE Add $i$ to the candidate set $\mathcal{C}$.
      \ENDIF
    \ENDFOR

    \STATE \textbf{Interaction Detection:}
    \FOR{$i \in \mathcal{C}$}
      \STATE Define geometry $S_i(t)$ for agent $i$, and $S_{\mathrm{target}}(t)$ for the target agent.
      \STATE \textbf{If} $\Bigl(\forall\,t:\;S_i(t)\cap S_{\mathrm{target}}(t)=\varnothing\Bigr) \,\wedge\, \bigl(\min_{t}d_i(t) \le \tau_{\mathrm{ignore}}\bigr)$:
      \quad label agent $i$ as $\texttt{nearby}$.
      \STATE \textbf{Else if} $\exists\,t:\;S_i(t)\cap S_{\mathrm{target}}(t)\neq\varnothing$:
      \quad compare the time steps of minimal distances to assign either $\texttt{overtaking}$ or $\texttt{yielding}$.
    \ENDFOR

    \STATE \textbf{Label Assignment:}
    \STATE Convert each final label \{\texttt{ignored}, \texttt{nearby}, \texttt{overtaking}, \texttt{yielding}\} 
           into a one-hot vector $H_i \in \mathbb{R}^4$.

    \RETURN $H$
\end{algorithmic}
\end{algorithm}

\section{Per-Category Marginal and Joint Prediction}
\label{sec:per-category}
As shown in \ref{tab:details_marginal_prediction_on_WOMD} and \ref{tab:joint_prediction_on_WOMD}, we present a comprehensive per-category evaluation of our framework's performance on both single-agent (marginal) and multi-agent (joint) motion prediction tasks using the Waymo Open Motion Dataset\cite{waymo_motion_prediction}.

\section{Additional Qualitative Results}
\label{sec:Appendix Qualitative Results}
\begin{figure*}[!t]
    \centering
    \includegraphics[width=16cm]{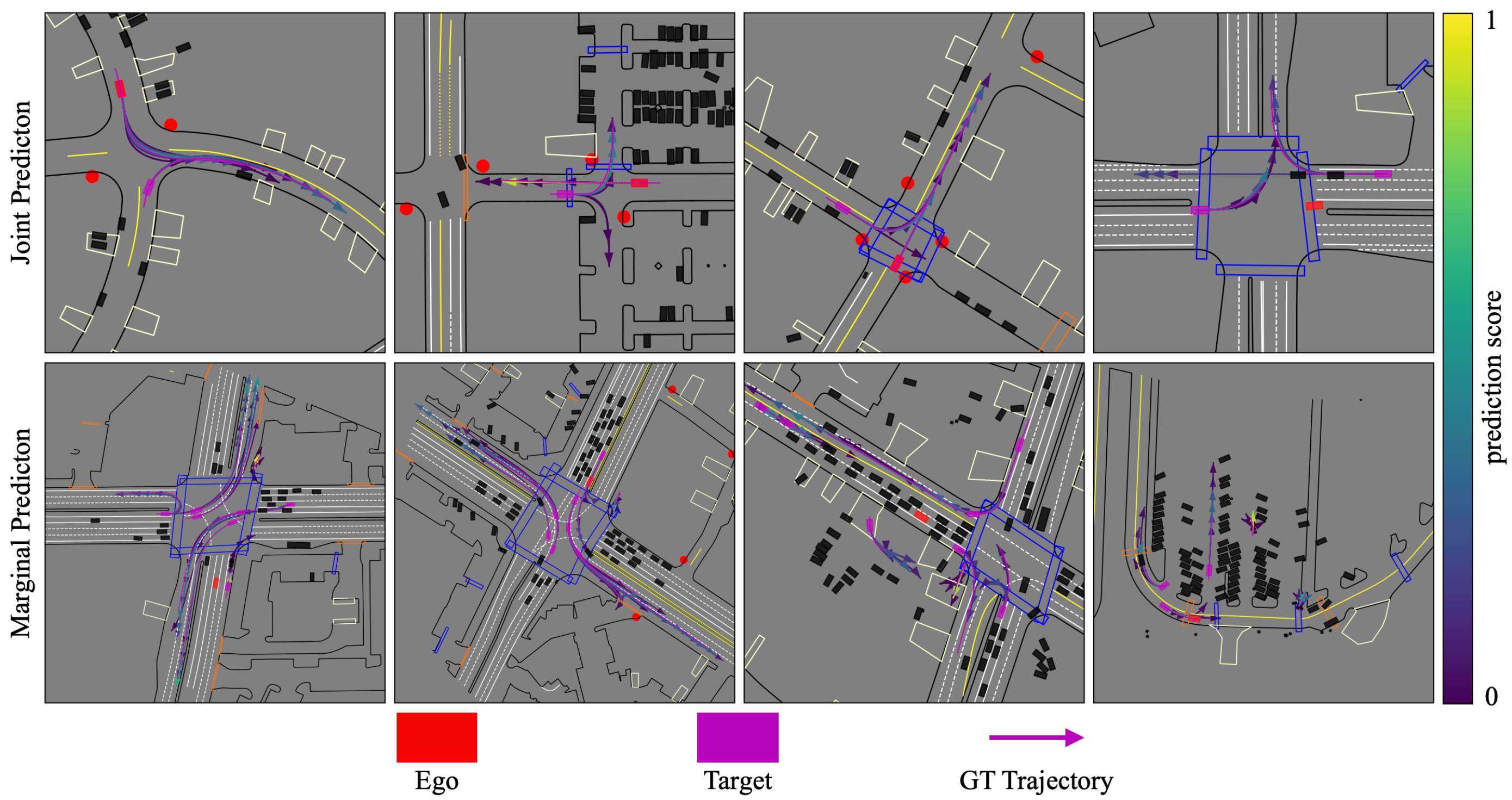} 
    \caption{Qualitative Results of IMPACT for Joint and Marginal Prediction on the WOMD Validation Dataset.}
    \label{fig:qualitative_marginal_joint}
    \vspace{-1em}
\end{figure*}

\textbf{Additional prediction results}.  Figure \ref{fig:qualitative_marginal_joint} provides qualitative prediction results under both joint and marginal settings.\\
\textbf{Visualization results of multimodal behavior intention prediction}. Figure \ref{fig:yoi} and \ref{fig:mode} provide detailed forecast visualization of multimodal trajectories and behavior intention labels.\\
\textbf{Behavior intention labels comparison between BeTop and IMPACT}. Figure \ref{fig:comparebetop} provides ground truth behavior intention labels comparison between BeTop and IMPACT.\\
\textbf{Visualization results of multimodal vectorized occupancy}. Figure \ref{fig:predocc} and \ref{fig:compocc} provide detailed forecast visualization of multimodal trajectories and vectorized occupancy prediction.

\section{Deployment Detail}
We re-trained our method on a large-scale private dataset comprising 8.7 million clips for training and 1 million clips for validation. Each clip includes one second of historical data and eight seconds of future data. Our method is integrated into an end-to-end planning framework, in which it consumes upstream perception outputs—such as object detection, tracking, and map reconstruction—and generates trajectory predictions for the downstream planning module. We have successfully deployed our system on real vehicles, and a demonstration video is provided in the supplementary materials.

\section{Notations}
As shown in \ref{tab:notations}, we provide a lookup table for notations in the paper.

\begin{table*}
\centering
\caption{Detailed performance of marginal predictions on WOMD Motion Leaderboard.}
\label{tab:details margin}
\begin{tabular}{c|c|cccccc}
    \hline
      \rowcolor[rgb]{0.863,0.863,0.863} Our Model ~& Category &Soft mAP $\uparrow$ & mAP $\uparrow$ & minADE $\downarrow$ & minFDE $\downarrow$ & Miss rate $\downarrow$ & Overlap Rate $\downarrow$\\
    \hline
    \\[-1em]
    \multirow{4}{*}{IMPACT} & Vehicle & 0.5040 & 0.4897 & 0.6696 & 1.3672 & 0.1122 & 0.0430 \\
    & Pedestrian & 0.4963 & 0.4866 & 0.3394 & 0.7009 & 0.0644 & 0.2651\\
    & Cyclist & 0.4159 & 0.4063 & 0.6832 & 1.3940 & 0.1662 & 0.0684\\
    & \textbf{Avg} & 0.4721 & 0.4609 & 0.5641 & 1.1540 & 0.1143 & 0.1255\\
    \hline
    \\[-1em]
    \multirow{4}{*}{\begin{tabular}[c]{@{}c@{}} IMPACT \\ (Ensemble)\end{tabular}} & Vehicle & 0.5035 & 0.4766 & 0.6668 & 1.3584 & 0.1111 & 0.0398\\
    & Pedestrian & 0.4996 & 0.4887 & 0.3386 & 0.7011 & 0.0637 & 0.2664\\
    & Cyclist & 0.4371 & 0.4142 & 0.6635 & 1.3290 & 0.1513 & 0.0713\\
    & \textbf{Avg} & 0.4801 & 0.4598 & 0.5563 & 1.1295 & 0.1087 & 0.1258\\
    \hline
    \\[-1em]
    \multirow{4}{*}{\begin{tabular}[c]{@{}c@{}} IMPACT \\ (e2e)\end{tabular}} & Vehicle & 0.4859 & 0.4657 & 0.6157& 1.2309 & 0.1194 & 0.0464\\
    & Pedestrian & 0.4599 & 0.4389 & 0.3114 & 0.6360 & 0.0716 & 0.2668\\
    & Cyclist & 0.3842 & 0.3712 & 0.6419 & 1.2822 & 0.1911 & 0.0774\\
    & \textbf{Avg} & 0.4434 & 0.4253 & 0.5230 & 1.0497 & 0.1274 & 0.1302\\
    \hline
    
\end{tabular}
\end{table*}

\begin{table*}
\centering
\caption{Marginal predictions on WOMD Motion Leaderboard}
\label{tab:details_marginal_prediction_on_WOMD}
\begin{tabular}{c|c|ccc|cc} 
\hline
\rowcolor[rgb]{0.863,0.863,0.863} 
\textbf{Category} & \textbf{Method} & \textbf{minADE} & \textbf{minFDE} & \textbf{Miss Rate} & \textbf{mAP} & \textbf{Soft mAP} \\ 
\hline
\multirow{4}{*}{Vehicle} 
    & MTR\cite{shi2023MTR}      & 0.7642 & 1.5257 & 0.1514 & 0.4494 & 0.4590 \\
    & ModeSeq\cite{zhou2024modeseq}      & 0.678 & 1.4046 & 0.1129 & \textbf{0.5095} & \textbf{0.5181} \\
    & BeTopNet\cite{liu2024reasoningmultiagentbehavioraltopology} & 0.6814 & 1.3888 & 0.1172 & 0.4860 & 0.4995 \\
    & IMPACT & \textbf{0.6668} & \textbf{1.3584} & \textbf{0.1111} &0.4766 &0.5035 \\ 
\hline
\multirow{4}{*}{Pedestrian} 
    & MTR\cite{shi2023MTR}      & 0.3486 & 0.7270 & 0.0753 & 0.4331 & 0.4409 \\
    & ModeSeq\cite{zhou2024modeseq}      & 0.3407 & 0.7117 & 0.0776 & 0.4709 & 0.4781 \\
    & BeTopNet\cite{liu2024reasoningmultiagentbehavioraltopology} & 0.3451 & 0.7142 & 0.0668 & 0.4777 & 0.4875 \\
    & IMPACT & \textbf{0.3386} & \textbf{0.7011} & \textbf{0.0637} &\textbf{0.4887} &\textbf{0.4996} \\
\hline
\multirow{4}{*}{Cyclist} 
    & MTR\cite{shi2023MTR}      & 0.7022 & 1.4093 & 0.1786 & 0.3561 & 0.3650 \\
    & ModeSeq\cite{zhou2024modeseq}      & 0.6852 & 1.4136 & 0.1706 & \textbf{0.4190} & 0.4248 \\
    & BeTopNet\cite{liu2024reasoningmultiagentbehavioraltopology} & 0.6905 & 1.3975 & 0.1688 & 0.4060 & 0.4163 \\
    & IMPACT & \textbf{0.6635} & \textbf{1.3290} & \textbf{0.1513} &0.4142 &\textbf{0.4371} \\
\hline
\end{tabular}
\end{table*} 

\begin{table*}
\centering
\caption{Joint predictions on WOMD Interaction Leaderboard}
\label{details_joint}

\begin{tabular}{c|c|ccc|cc} 
\hline
\rowcolor[rgb]{0.863,0.863,0.863} 
\textbf{Category} & \textbf{Method} & \textbf{minADE} & \textbf{minFDE} & \textbf{Miss Rate} & \textbf{mAP} & \textbf{Soft mAP} \\ 
\hline
\multirow{4}{*}{Vehicle} 
    & GameFormer\cite{huang2023gameformergametheoreticmodelinglearning} & 1.0499 & 2.4044 & 0.4321 & 0.2469 & 0.2564 \\
    & BeTopNet\cite{liu2024reasoningmultiagentbehavioraltopology} & \textbf{1.0216} & \textbf{2.3970} & 0.3738 & 0.3374 & 0.3308 \\
    & IMPACT & 1.0267 & 2.4162 & \textbf{0.3713} &\textbf{0.3380} &\textbf{0.3450} \\ 
\hline
\multirow{4}{*}{Pedestrian} 
    & GameFormer\cite{huang2023gameformergametheoreticmodelinglearning} & 0.7978 & \textbf{1.8195} & 0.4713 & 0.1962 & 0.2014 \\
    & BeTopNet\cite{liu2024reasoningmultiagentbehavioraltopology} & 0.7862 & 2.3970 & \textbf{0.3738} & \textbf{0.3374} & \textbf{0.3308} \\
    & IMPACT & \textbf{0.7834} & 1.8198 & 0.4005 &0.2849 &0.2917 \\ 
\hline
\multirow{4}{*}{Cyclist} 
    & GameFormer\cite{huang2023gameformergametheoreticmodelinglearning} & \textbf{1.0686} & \textbf{2.4199} & 0.5765 & 0.1367 & 0.1338 \\
    & BeTopNet\cite{liu2024reasoningmultiagentbehavioraltopology} & 1.1155 & 2.5850 & 0.5253 & 0.1717 & 0.1756 \\
    & IMPACT & 1.1113 & 2.5841 & \textbf{0.5231} &\textbf{0.1748} &\textbf{0.1788} \\ 
\hline
\end{tabular}
\label{tab:joint_prediction_on_WOMD}
\end{table*}



\small {
\begin{table*}[!h]
\centering
\caption{List of important notations in this work}
\label{tab:notations}
\begin{tabular}{lll}
\toprule
\textbf{Symbol} & \textbf{Meaning} & \textbf{Dimension} \\
\midrule
$N_{a}$ & Number of agents in the scene & Scalar \\
$N_{l}$ & Number of polylines in the map & Scalar \\
$N_{p}$ & Number of points in each polyline & Scalar \\
$T_{p}$ & Number of past timesteps (observation horizon) & Scalar \\
$T_{f}$ & Number of future timesteps (prediction horizon) & Scalar \\
$K$ & Number of future modes (prediction branches) & Scalar \\
$M$ & Number of query content features/modes in the decoder & Scalar \\
$D$ & Feature dimension in the latent space & Scalar \\
$\mathcal{A} = \{a_{1}, \dots, a_{N_a}\}$ & Set of all agent trajectories (history) & $\mathbb{R}^{N_a \times T_p \times F_1}$ \\
$\mathcal{L} = \{l_{1}, \dots, l_{N_l}\}$ & Set of all map polylines & $\mathbb{R}^{N_l \times N_p \times F_2}$ \\
$\mathcal{R} = \{r_{1}, \dots, r_{N_l}\}$ & Historical relative movement features & $\mathbb{R}^{N_l \times T_p \times F_3}$ \\
$y_{i}^{k}$ & $k$-th predicted future trajectory for agent $i$ & $\mathbb{R}^{T_f \times 2}$ \\
$s_{i}^{k}$ & Confidence score for $k$-th future trajectory & Scalar \\
$\mathcal{H}^k = \{h_{1}^k, \dots, h_{N_a}^k\}$ & behavioral intentions under mode $k$ & $\mathbb{R}^{N_a \times 4}$ \\
$\mathcal{O}^k = \{o_{1}^k, \dots, o_{N_l}^k\}$ & Occupancy probabilities under mode $k$ & $\mathbb{R}^{N_l \times 1}$ \\
$\mathcal{A}_k^{\mathrm{sel}}$ & Top-$m$ selected agents under mode $k$ & $\mathbb{R}^{m \times D}$ \\
$\mathcal{L}_k^{\mathrm{sel}}$ & Top-$n$ selected polylines under mode $k$ & $\mathbb{R}^{n \times D}$ \\
$\phi(\cdot), \varphi(\cdot)$ & Utility functions deriving scalar scores from vectors & -- \\
$Q$ & Query content features in the decoder & $\mathbb{R}^{M \times D}$ \\
$\hat{\mathrm{O}}$ & Predicted vectorized occupancy & $\mathbb{R}^{K \times N_l \times 1}$ \\
$\mathrm{O}^{*}$ & Ground-truth vectorized occupancy & $\mathbb{R}^{N_l \times 1}$ \\
$\hat{\mathrm{H}}$ & Predicted behavioral intentions & $\mathbb{R}^{K \times N_a \times 4}$ \\
$\mathrm{H}^{*}$ & Ground-truth behavioral intentions & $\mathbb{R}^{N_a \times 4}$ \\
$\hat{\mathrm{Y}}$ & Predicted Future Gaussian components & $\mathbb{R}^{K \times T_{f} \times 5}$ \\
$\mathrm{Y}^{*}$ & Ground-truth Future Trajectories & $\mathbb{R}^{T_{f} \times 2}$ \\
$\hat{\mathrm{S}}$ & Predicted scores & $\mathbb{R}^{K}$ \\
\bottomrule
\end{tabular}
\end{table*}
}

\end{document}